\documentclass[11pt]{article}
\usepackage[preprint]{acl}
\usepackage{times}
\usepackage{latexsym}
\usepackage[T1]{fontenc}
\usepackage[utf8]{inputenc}
\usepackage{microtype}
\usepackage{inconsolata}
\usepackage{graphicx}
\usepackage{amsmath}
\usepackage{amssymb}
\usepackage{booktabs}
\usepackage{multirow}
\usepackage{longtable}
\usepackage{float}
\usepackage{xspace}
\usepackage{pifont}
\usepackage{xcolor}

\title{Alignment Makes Language Models Normative, Not Descriptive}

\author{Eilam Shapira \and Moshe Tennenholtz \and Roi Reichart \\
Technion -- Israel Institute of Technology}

\begin{document}
\maketitle

\begin{figure*}[t]
\centering
\includegraphics[width=\textwidth]{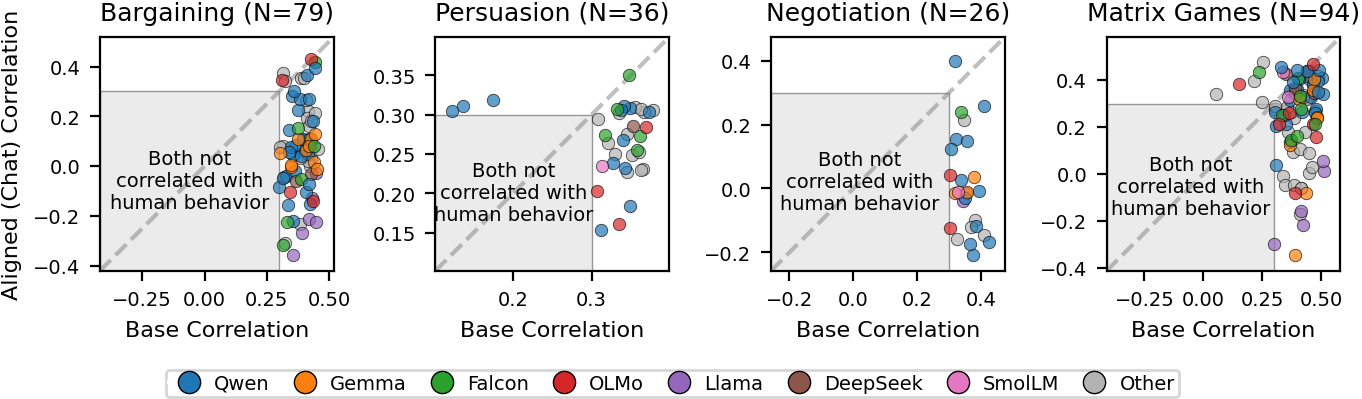}
\caption{Pearson correlations of base models and human decisions (x-axis) vs.\ aligned models and human decisions (y-axis) across four game families. Each point is a same-provider pair evaluated in its native format (standard prompt for base, chat template for aligned). \textbf{Points below the diagonal indicate base advantage.} The shaded region marks pairs where both models correlate below 0.3 with human behavior. Base models win 75:4 in bargaining, 32:4 in persuasion, 25:1 in negotiation, and 81:13 in matrix games, for an overall ratio of 9.7:1 (213 vs.\ 22, $p < 10^{-40}$).}
\label{fig:main}
\end{figure*}

\begin{abstract}
Post-training alignment optimizes language models to match human preference signals, but this objective is not equivalent to modeling observed human behavior. We compare 120 base–aligned model pairs on more than 10,000 real human decisions in multi-round strategic games—bargaining, persuasion, negotiation, and repeated matrix games. In these settings, base models outperform their aligned counterparts in predicting human choices by nearly 10:1, robustly across model families, prompt formulations, and game configurations.
This pattern reverses, however, in settings where human behavior is more likely to follow normative predictions: aligned models dominate on one-shot textbook games across all 12 types tested and on non-strategic lottery choices — and even within the multi-round games themselves, at round one, before interaction history develops. This boundary-condition pattern suggests that alignment induces a normative bias: it improves prediction when human behavior is relatively well captured by normative solutions, but hurts prediction in multi-round strategic settings, where behavior is shaped by descriptive dynamics such as reciprocity, retaliation, and history-dependent adaptation. These results reveal a fundamental trade-off between optimizing models for human use and using them as proxies for human behavior.
\end{abstract}

\section{Introduction}
\label{sec:intro}

Large language models (LLMs) are increasingly used as proxies for human behavior \citep{horton2023homo,aher2023using,binz2023using,argyle2023out,santurkar2023whose,hewitt2024predicting,suh2025subpop}. They replicate classic experimental findings from psychology and economics, approximate subgroup opinion distributions when conditioned on demographic backstories, and predict survey experiment outcomes. The approach extends to strategic settings: LLMs can predict human decisions in language-based persuasion games, outperforming models trained on human data alone \citep{shapira2024llm}, and capture cooperation patterns in repeated social dilemmas \citep{akata2025playing, mei2024turing}.

Yet nearly all of this work uses aligned models, treating alignment as either neutral or beneficial for behavioral prediction. This assumption deserves scrutiny. Alignment via RLHF \citep{ouyang2022training} or DPO \citep{rafailov2023direct} optimizes models for responses that human evaluators \emph{approve of}--cooperative, fair, and socially appropriate. But human behavior in strategic settings is often none of these: people bluff, retaliate, and deviate from approved patterns \citep{capraro2025optimistic,bauer2025gpt4}. If alignment narrows the model's behavioral distribution toward such responses \citep{kirk2024understanding,cao2025entropy,chen2025kl}, it creates a \emph{normative bias}--the model learns to predict behavior that people \emph{endorse} rather than behavior they \emph{exhibit}. The distinction between normative theories (how people should act) and descriptive accounts (how people actually act) is foundational in the social and behavioral sciences \citep{camerer2004cognitive}.

This predicts that aligned models should predict human behavior well in settings where that behavior is relatively simple and well-described by normative theory, but poorly where behavior is complex and shaped by interaction history. Multi-round strategic games--where decisions depend on accumulated experience with a specific opponent--provide a natural test case for the descriptive end: behavior there is driven by reciprocity, retaliation, and reputation dynamics. One-shot decisions over well-studied game structures or simple lotteries provide a contrasting case where normative predictions may be more accurate.

We test this hypothesis by comparing 120 same-provider base--aligned\footnote{Throughout, \emph{aligned} denotes models that have undergone post-training optimization beyond next-token prediction--typically supervised fine-tuning combined with preference optimization via RLHF or DPO; \emph{base} denotes the pre-alignment checkpoint.} model pairs from 23 families (see Appendix~\ref{app:models}) on predicting 10,050 real human decisions across four families of multi-round strategic games: bargaining, persuasion, negotiation, and repeated matrix games (Prisoner's Dilemma and Battle of the Sexes). By restricting to same-provider pairs, each comparison directly isolates the effect of alignment. Each model is evaluated in its native format: standard text completion for base models, chat-templated input for aligned models.

The results are consistent with the hypothesis. In multi-round games, base models outperform their aligned counterparts by a ratio of 9.7:1 (213 vs.\ 22 wins, $p < 10^{-40}$), with each game family individually significant ($p < 10^{-6}$). The effect holds across all 23 model families, 10 prompt formulations, and all game configuration parameters, and grows with model scale.

The hypothesis also predicts where the base advantage should \emph{not} hold: in simpler settings without multi-round history, normative predictions may suffice, and alignment should help rather than hurt. We test two such boundary conditions--one-shot $2\times2$ matrix games and non-strategic binary lotteries--and find that the advantage reverses in both. Aligned models win 4.1:1 on one-shot games ($p < 10^{-6}$), consistently across all 12 game types, and 2.2:1 on lotteries ($p < 10^{-3}$). In the one-shot games, aligned models' predictions are closer to Nash equilibrium--which itself correlates with human behavior in these settings ($r = 0.62$)--consistent with alignment shifting predictions toward normative patterns. The same reversal appears within multi-round games at round one, before interaction history develops, but disappears as history accumulates.

\section{Related Work}

\subsection{LLMs as Human Behavioral Proxies}

A growing literature treats LLMs as behavioral models of humans--\emph{homo silicus} \citep{horton2023homo}--capable of replicating experimental findings \citep{aher2023using}, approximating subgroup opinions \citep{argyle2023out}, and predicting treatment effects \citep{hewitt2024predicting}. Nearly all of this work uses aligned models, implicitly assuming that alignment is neutral for behavioral fidelity. Yet several findings challenge this assumption: RLHF collapses opinion diversity toward specific groups \citep{santurkar2023whose}, instruction tuning introduces cognitive biases absent in base models \citep{itzhak2024instructed}, LLMs over-predict normatively rational behavior \citep{liu2025llms}, and RLHF-tuned models fail to mirror human response biases \citep{tjuatja2024llms}. Most directly, \citet{suh2025subpop} found that aligned models are dramatically worse than base models at zero-shot opinion prediction. These results suggest that alignment distorts behavioral representations--but the evidence comes from opinions and individual judgments. Whether the pattern extends to \emph{multi-round strategic interactions}, where behavior is shaped by history and reciprocity, remains untested.

\subsection{The Alignment Tax}

Alignment can degrade capabilities beyond helpfulness, a phenomenon termed the ``alignment tax.'' Base models outperform aligned variants on reasoning benchmarks \citep{munjal2026base}, and calibration deteriorates across the tuning pipeline \citep{kadavath2022language, zhu2023calibration}. More fundamentally, alignment narrows the model's output distribution: RLHF significantly reduces output diversity \citep{kirk2024understanding}, and the standard KL-regularized RL framework can only specify unimodal targets, making diversity collapse a built-in feature rather than an implementation failure \citep{chen2025kl, korbak2022rl, xiao2024bias}. These results establish \emph{that} alignment narrows distributions and \emph{why}, but measure the cost in generation quality and benchmark scores--not in behavioral prediction fidelity. Whether distributional narrowing degrades a model's ability to predict the full range of human strategic behavior has not been tested directly.

\subsection{LLMs in Strategic Games}

Prior work studies how LLMs \emph{play} games \citep{capraro2025optimistic, akata2025playing, mei2024turing} or serve as available strategies \citep{shapira2026poisoned}, but play and prediction are fundamentally different: a model at Nash equilibrium would poorly predict actual human behavior, which systematically deviates from equilibrium. We study \emph{prediction}--whether a model's token probabilities match human choice distributions--using logprob extraction rather than generation, enabling direct base-vs-aligned comparison on identical inputs.

Predicting human strategic behavior has traditionally relied on parametric models from behavioral game theory \citep{mckelvey1995quantal, mckelvey1998quantal, nagel1995unraveling, stahl1995players, camerer2004cognitive, camerer1999experience}. \citet{zhu2025predicting} showed that ML models trained on large human datasets capture structure beyond these baselines, and \citet{shapira2024llm, shapira2024glee, shapira2025prediction} demonstrated that LLMs can predict human decisions in language-based games--but used only aligned models, leaving open whether the pre-alignment checkpoint might predict better. We address this gap with the first systematic base-vs-aligned comparison across 120 same-provider pairs and four game families.

\section{Experimental Setup}

\subsection{Game Families and Human Data}

We evaluate on four families of strategic games that vary in information structure, decision complexity, and interaction length.

\paragraph{Bargaining.} An alternating-offers bargaining game based on the model of \citet{rubinstein1982perfect}. Alice and Bob take turns proposing how to divide a sum of money; the other player accepts or rejects. Each player has a per-round discount factor ($\delta_1$, $\delta_2$) representing value loss over time, framed to participants as ``inflation.'' Proposals are accompanied by optional free-text messages. If no agreement is reached within the allotted rounds, both players receive nothing. The human participant plays one role and makes binary accept/reject decisions at each of their turns. This family contains 1,788 human decisions.

\paragraph{Persuasion.} A repeated cheap talk game \citep{crawford1982strategic} played over 20 rounds. Each round, a seller observes whether a product is high- or low-quality (drawn independently) and sends a message to a buyer, who then decides whether to purchase at a fixed price. The seller profits from every sale regardless of quality, creating a credibility problem: the unique stage-game equilibrium is babbling (uninformative messages). Over repeated rounds, however, reputation dynamics emerge as buyers observe the seller's track record. The buyer role comes in two variants: a \emph{long-living} buyer who observes the full history, and \emph{myopic} buyers who see only aggregate statistics. Human participants play the buyer role and make binary yes/no decisions. This family contains 3,180 human decisions.

\paragraph{Negotiation.} A bilateral price negotiation in which a seller and buyer alternate price proposals for an indivisible good. Each player has a private valuation: the seller values the good at $V_A$ and the buyer at $V_B$ (parameterized as multiples of a base price). At each decision point, the responding player can accept the current price, reject it (passing the initiative to the other side), or exercise an outside option--transacting with an alternative partner ``John'' at their own valuation, guaranteeing zero surplus but ending the negotiation.\footnote{The outside option was introduced in GLEE to provide a credible disagreement point; without it, rejection merely delays the game, incentivizing acceptance even at unfavorable prices. For evaluation we code both reject and DealWithJohn as~0 (non-accept), since both represent refusal of the current offer.} Human decisions are ternary: AcceptOffer, RejectOffer, or DealWithJohn. This family contains 1,182 human decisions.

These three families are drawn from the GLEE benchmark \citep{shapira2024glee}. In GLEE, human participants play interactively against LLM opponents through a web interface: each human takes one role in a game while an LLM plays the other, producing natural language dialogues with varied offers, arguments, and counteroffers. Participants were not informed that their opponent was an LLM; the interface presented the other player by name (e.g., ``Alice''), so human decisions were uncontaminated by knowledge of the opponent's nature. The resulting game transcripts contain decision points where humans chose among discrete actions within rich, multi-turn conversational contexts.

\paragraph{Repeated $2\times2$ Matrix Games.} We additionally evaluate on two repeated $2\times2$ games from \citet{akata2025playing}: the Prisoner's Dilemma (PD) and the Battle of the Sexes (BoS). In each, 195 human participants play 10 rounds against pre-computed opponent strategies derived from GPT-4, yielding 1,950 decisions per game (3,900 total). Participants were told they might face a human or an artificial agent; in fact, all played against LLMs, with debriefing provided afterward. In PD, participants choose to cooperate or defect; in BoS, they coordinate on one of two options with asymmetric preferences. Unlike the GLEE games, these are complete-information games with a known payoff matrix. We format these games using a multi-turn prompt structure, presenting the payoff matrix and round history as a structured dialogue.

Across all four families, our evaluation covers 10,050 human decisions per model, yielding over 2.4 million total predictions across all models and pairs.

\subsection{Prediction Method}

We frame human decision prediction as a token probability extraction task. For each human decision point in a game, we construct a prompt consisting of a system message describing the game rules and the participant's role, followed by the dialogue history up to the decision point. We then perform a single forward pass through the model and extract the log-probabilities assigned to each decision token (e.g., ``accept'' vs.\ ``reject'' for bargaining) from the model's next-token distribution at the final position.

We normalize the extracted probabilities to obtain a predicted decision distribution:

\begin{equation}
p_{\text{accept}} = \frac{p(\text{yes})}{\sum_{d} p(d)}
\end{equation}

where $d$ ranges over all decision tokens for a given family (two tokens for bargaining, persuasion, and matrix games; three for negotiation, which adds the outside-option token). The resulting $p_{\text{accept}} \in [0,1]$ captures the model's relative preference for the affirmative action, normalized away from non-decision tokens.

This method requires no text generation and no sampling--it is a deterministic extraction of the model's internal probability distribution over decision tokens, applicable to both base and aligned models without requiring different decoding strategies. The normalization is robust when decision tokens receive substantial probability mass; when they do not (i.e., the model distributes mass primarily to non-decision tokens), the normalized probabilities become unreliable. We therefore apply two pair-level filters per game family: a \emph{mass filter} excluding pairs where either model assigns less than 80\% average probability mass to decision tokens, and a \emph{minimum correlation filter} excluding pairs where \emph{both} models correlate below 0.3 with human decisions. Filters are applied independently per family; the base advantage is robust across threshold choices (see Appendix~\ref{app:filter_details}).
\label{sec:filtering}

\subsection{Prompt Variants}

We evaluate four prompt variants per model pair to disentangle the effects of model type (base vs.\ aligned) and prompt format. All variants append a partial JSON object (e.g., \texttt{\{"decision": "}) after the dialogue history, prompting the model to complete it with a decision token. The \emph{standard} format presents this directly as a text completion; the \emph{chat template} format additionally wraps the prompt in the formatting tokens expected by aligned models (e.g., \texttt{<|im\_start|>}, \texttt{[INST]}), structuring the input into system, user, and assistant roles.

The four variants cross model type with format: \textbf{Base (native)} uses standard format; \textbf{Aligned (native)} uses the model's chat template; \textbf{Base (chat)} applies the aligned partner's chat template to the base model; and \textbf{Aligned (plain)} uses standard format without chat template. Our \textbf{main comparison} pairs each model in its native format--base with standard, aligned with chat template--reflecting the most natural deployment condition. The two additional variants serve as controls: \emph{Base (chat)} tests whether applying the aligned model's chat template to its base counterpart can recover any aligned-model advantage, while \emph{Aligned (plain)} tests aligned models in a format they were not optimized for.

To test whether the base advantage depends on prompt wording, we evaluate 14 additional formulations spanning framing, persona, format, and structure modifications (see Appendix~\ref{app:variants}). Results are reported in Section~\ref{sec:prompt_robustness}.

\subsection{Boundary Condition Datasets}

We additionally evaluate on two datasets chosen to test the limits of the base advantage.

\paragraph{One-shot $2\times2$ matrix games.} We use a dataset of 2,416 procedurally generated one-shot $2\times2$ matrix games from \citet{zhu2025predicting}, spanning 12 game topologies with approximately 93,000 aggregated human decisions. Unlike our repeated matrix games, these are single-round decisions over well-studied game structures that are abundantly represented in LLM training data. We present games in counterbalanced format (swapping row labels to control for position bias). After filtering, 71 valid pairs remain.

\paragraph{Binary lottery choices.} We use the dataset of \citet{marantz2025binary}, comprising 1,001 binary lottery choice problems in which each of 28--31 participants chooses between two gambles specified by their outcomes and probabilities (e.g., ``\$10 with 60\% or \$2 otherwise'' vs.\ ``\$7 with 80\% or \$1 otherwise''). We present these using verbal descriptions of each lottery. After filtering, 90 valid same-provider pairs remain. These are non-strategic decisions--there is no opponent or interaction--allowing us to test whether the base advantage is specific to strategic reasoning or extends to individual decision-making under risk.

\subsection{Evaluation}

\paragraph{Primary metric.} We use Pearson correlation between the model's predicted probability ($p_{\text{accept}}$) and the ground-truth human behavior as our primary evaluation metric. In the four main game families (bargaining, persuasion, negotiation, repeated matrix games), each decision point has a unique dialogue history, so the correlation is computed at the level of individual decisions (coded as 1 for accept/yes/cooperate, 0 for reject/no/defect; in negotiation, both reject and DealWithJohn are coded as~0). In the boundary condition datasets (one-shot $2\times2$ games and lottery choices), the same problem is presented to multiple participants, yielding an empirical choice probability per problem; here, we correlate the model's predicted probability with this aggregate human choice rate. This reflects the data structure: multi-round games produce unique trajectories, while one-shot problems are repeated across participants.

\paragraph{Pairwise comparison.} For each base--aligned pair in a given game family, we compare the base model's Pearson correlation against the aligned model's Pearson correlation and record a ``base win'' or ``aligned win.'' We then aggregate win counts across all valid pairs.

\paragraph{Statistical tests.} We employ two complementary tests. A one-sided binomial test evaluates whether the observed majority (base or aligned) wins significantly more than 50\% of comparisons under the null hypothesis of equal performance; the test is always applied in the direction of the observed winner. As a complementary test that accounts for effect magnitudes, we also report the one-sided Wilcoxon signed-rank test on the Pearson correlation differences. All $p$-values reported in the text are binomial unless otherwise noted.

\section{Results}
\label{sec:results}

Figure~\ref{fig:main} visualizes the head-to-head comparison under our main pairing: each model in its native format (standard prompt for base, chat template for aligned). Base models win 213 of 235 valid comparisons across the four game families (9.7:1), with the advantage individually significant in every family ($p < 10^{-6}$). The advantage is consistent across all 23 model families. Among the seven largest, base wins the majority in every family: Qwen 82:15, Gemma 28:2, Falcon 21:6, Llama 17:0, OLMo 16:3, DeepSeek 8:4, and SmolLM 5:3. Even the families closest to parity never show a consistent aligned-model advantage across game types. Full per-pair results for all six datasets are reported in Appendix~\ref{app:per_pair}.

\paragraph{Ruling out prompt-format confounds.} A natural objection is that base models benefit from plain-text format while aligned models are hampered by their chat template. Two controls rule this out: when both models receive identical plain-text prompts, base models still win 5.0:1 ($p < 10^{-34}$); when both receive the aligned model's chat template--a format the base model was never trained on--base models still win 5.3:1. The advantage resides in the model weights, not in the prompt format.

\paragraph{Prompt formulation robustness.}
\label{sec:prompt_robustness}

We evaluate 14 prompt formulations organized into four clusters--framing (3 variants modifying task description, e.g., ``predict what a human would do'' or casting the model as an external observer), persona (5 variants assigning behavioral roles: naive, expert, fairness-oriented, selfish, and emotional), format (3 variants stripping structured formatting), and structure (2 variants altering prompt organization)--plus the baseline. Of these, 10 produce sufficient data for evaluation; the natural language and simplified format variants yield catastrophically low decision token mass for base models, indicating reliance on structured formatting. Across the 10 testable variants and two GLEE game families (bargaining and negotiation), base models win 959 of 1,003 comparisons (95.6\%, $p < 10^{-200}$), with every variant individually reaching $p < 0.01$ (Table~\ref{tab:prompt_variants}).

\begin{table}[t]
\centering
\footnotesize
\caption{Base (B) vs.\ aligned (A) win counts by prompt variant. Each variant pairs base predictions against the matching aligned chat variant. $p$: one-sided binomial test. Persuasion yields no valid pairs for non-standard variants.}
\label{tab:prompt_variants}
\resizebox{\columnwidth}{!}{%
\begin{tabular}{@{}llrrl@{}}
\toprule
\textbf{Cluster} & \textbf{Variant} & \textbf{B} & \textbf{A} & \textbf{$p$} \\
\midrule
Baseline  & Standard          & 101 & 5  & $1.3\!\times\!10^{-24}$ \\
Framing   & Predict human     & 105 & 3  & $6.5\!\times\!10^{-28}$ \\
          & Observer          & 92  & 5  & $4.3\!\times\!10^{-22}$ \\
          & Reversed roles    & 99  & 2  & $2.1\!\times\!10^{-27}$ \\
Persona   & Naive             & 104 & 5  & $1.9\!\times\!10^{-25}$ \\
          & Expert            & 101 & 5  & $1.3\!\times\!10^{-24}$ \\
          & Fairness          & 98  & 4  & $8.7\!\times\!10^{-25}$ \\
          & Selfish           & 100 & 6  & $2.2\!\times\!10^{-23}$ \\
          & Emotional         & 102 & 4  & $6.4\!\times\!10^{-26}$ \\
Structure & Preamble reversed & 57  & 5  & $1.5\!\times\!10^{-12}$ \\
\midrule
\textbf{Overall} &           & \textbf{959} & \textbf{44} & $< 10^{-200}$ \\
\bottomrule
\end{tabular}%
}
\end{table}
Framing, persona, and baseline variants all yield 92--97\% base win rates; even ``selfish'' (94.3\%) or ``observer'' (94.8\%) variants do not close the gap. Base models require structured formatting to produce valid decision tokens, but given such structure, the advantage is robust.

\paragraph{Game configuration robustness.}

Each GLEE family is parameterized along multiple dimensions (6 for bargaining, 6 for persuasion, 6 for negotiation; see Appendix~\ref{app:configs} for the full parameter space and per-value win counts). The base advantage holds across every parameter value in every family. In persuasion, the advantage is notably stronger when the seller knows product quality (14.5:1) than when uninformed (2.3:1), suggesting base models better capture strategic information use. The sole exception is bargaining with discount factor $\delta_1 = 0.8$, where the advantage narrows to near parity (10:7, $p = 0.31$).

\paragraph{Round-by-round dynamics.}
\label{sec:round_by_round}

In round~1--before any multi-round dynamics develop--aligned models actually win in bargaining (61:32), negotiation (39:33), and persuasion (30:23). The advantage reverses from round~2 onward (bargaining: 82:4, negotiation: 56:1, persuasion: 31:8).\footnote{Per-round analysis for repeated matrix games is not meaningful because round~1 contains only two unique decision contexts (one PD, one BoS), yielding insufficient variation for correlation.} This within-game transition mirrors the between-dataset contrast with one-shot games (Section~\ref{sec:boundary}), suggesting that the accumulation of history-dependent dynamics--not the game structure itself--drives the base advantage.

\paragraph{Size scaling.}

\begin{figure}[t]
\centering
\includegraphics[width=\columnwidth]{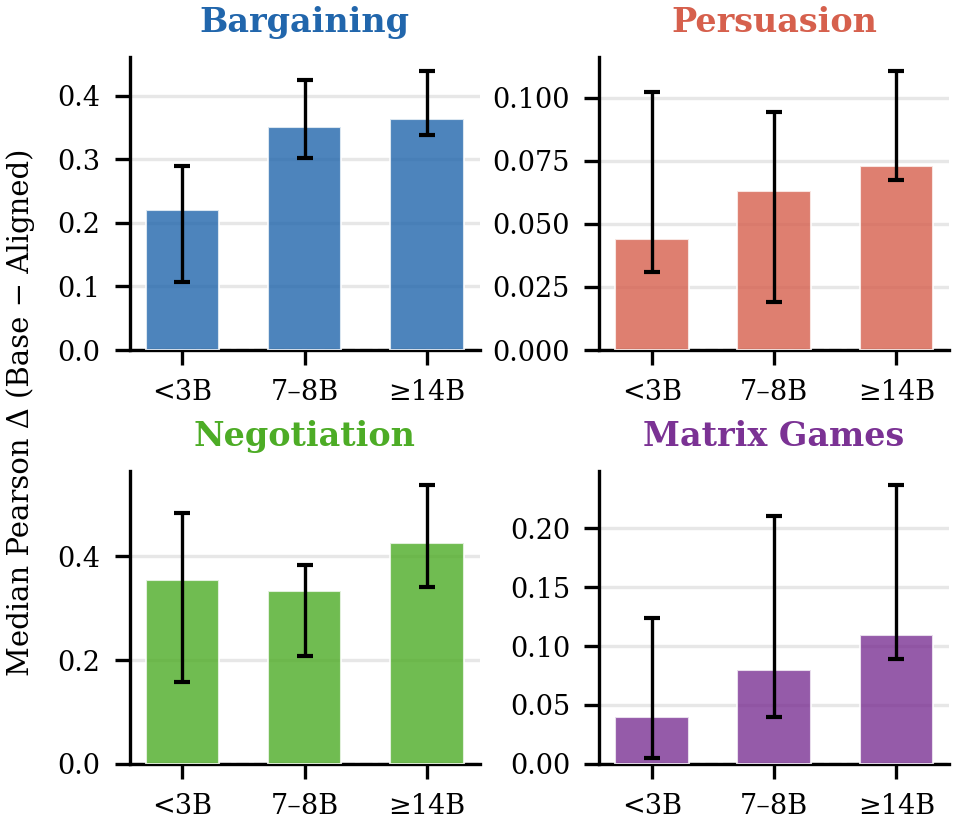}
\caption{Median Pearson correlation difference (base minus aligned) by model size, with 95\% bootstrap confidence intervals (5,000 resamples, percentile method). The base advantage is positive across all size bins and grows with scale.}
\label{fig:size}
\end{figure}

If the base advantage reflects richer pre-training representations that alignment shifts, it should grow with model scale. Figure~\ref{fig:size} confirms this: bargaining shows the clearest trend, from +0.22 at $<$3B to +0.36 at $\geq$14B; negotiation rises from +0.35 to +0.43; matrix games grow from +0.04 to +0.11.

\section{Boundary Conditions}
\label{sec:boundary}

The round-by-round analysis (Section~\ref{sec:round_by_round}) offers a clue about the limits of the base advantage: aligned models win at round~1, before interaction history develops, then lose as history accumulates. If the absence of multi-round history is what enables the aligned-model advantage, then it should reappear in settings that are inherently one-shot. We test this with two boundary conditions: one-shot matrix games (same strategic structure, no repeated interaction) and non-strategic lotteries (no opponent, no interaction). In both cases, the advantage reverses.

\begin{table}[t]
\centering
\small
\caption{Base vs.\ aligned wins on one-shot $2\times2$ games by game type ($N=71$ pairs). $p$: one-sided binomial test.}
\label{tab:oneshot}
\resizebox{\columnwidth}{!}{%
\begin{tabular}{@{}lrrll@{}}
\toprule
\textbf{Game Type} & \textbf{Base} & \textbf{Al.} & \textbf{Ratio} & \textbf{$p$ (binomial)} \\
\midrule
harmony     & 12 & 59 & 4.9:1 Al.  & $6.7 \times 10^{-9}$  \\
concord     & 9  & 62 & 6.9:1 Al.  & $3.7 \times 10^{-11}$ \\
peace       & 14 & 57 & 4.1:1 Al.  & $1.3 \times 10^{-7}$  \\
safecoord   & 15 & 56 & 3.7:1 Al.  & $5.2 \times 10^{-7}$  \\
assurance   & 15 & 56 & 3.7:1 Al.  & $5.2 \times 10^{-7}$  \\
dilemma     & 14 & 57 & 4.1:1 Al.  & $1.3 \times 10^{-7}$  \\
deadlock    & 16 & 55 & 3.4:1 Al.  & $1.9 \times 10^{-6}$  \\
chicken     & 23 & 48 & 2.1:1 Al.  & $2.0 \times 10^{-3}$  \\
staghunt    & 11 & 60 & 5.5:1 Al.  & $1.3 \times 10^{-9}$  \\
hero        & 23 & 48 & 2.1:1 Al.  & $2.0 \times 10^{-3}$  \\
leader      & 20 & 51 & 2.5:1 Al.  & $1.5 \times 10^{-4}$  \\
compromise  & 19 & 52 & 2.7:1 Al.  & $5.6 \times 10^{-5}$  \\
\midrule
\textbf{Per-pair} & \textbf{14} & \textbf{57} & \textbf{4.1:1 Al.}& $1.3 \times 10^{-7}$ \\
\bottomrule
\end{tabular}%
}
\end{table}

We evaluate 71 same-provider pairs on the one-shot $2\times2$ matrix game benchmark of \citet{zhu2025predicting}, comprising 2,416 procedurally generated games with approximately 93,000 aggregated human decisions spanning 12 game types. The results reverse: aligned models win 57 comparisons to base models' 14 (4.1:1 aligned-model advantage, $p < 10^{-6}$). The aligned-model advantage is universal across all 12 types (See Table~\ref{tab:oneshot}).
The contrast with repeated games is notable: when the same strategic structures are played over 10 rounds, base models win 6.2:1 (81:13). Two non-exclusive factors likely contribute: one-shot games are canonical objects abundantly represented in training corpora, where alignment may reinforce textbook-correct response patterns; and humans in isolated one-shot decisions may themselves behave closer to normative predictions--without the opportunity for reputation building, retaliatory spirals, or endgame effects, there is less room for the kind of behavioral divergence from norms that base models capture better. In repeated play, both factors reverse: the multi-round dynamics are novel rather than canonical, and humans adopt history-dependent strategies that diverge from textbook predictions.

We can quantify the normative alignment directly. For each one-shot game, we compute the mixed-strategy Nash equilibrium probability.\footnote{For games with multiple pure equilibria (33\% of the dataset), we use the unique mixed-strategy NE, which provides a single prediction per game without requiring an equilibrium selection assumption. At the population level, if different participants coordinate on different pure equilibria, aggregate choice frequencies converge toward the mixed NE prediction.} Human aggregate choices correlate with NE predictions ($r = 0.616$), suggesting that behavior in these simple games is reasonably well-described by equilibrium theory. Aligned models are systematically more NE-aligned than base models (mean $r = 0.41$ vs.\ $0.28$; aligned closer in 59 of 76 filtered pairs, $p < 10^{-6}$). This is consistent with alignment shifting predictions toward normative patterns--a shift that helps in settings where human behavior happens to follow such patterns.

We also evaluate on the lottery dataset of \citet{marantz2025binary}, comprising 1,001 binary choice problems with no strategic interaction. Among 90 same-provider pairs, aligned models win 62:28 (2.2:1, $p = 2.19 \times 10^{-4}$). Alignment helps with individual, non-interactive decisions, where understanding the decision structure and following instructions aligns with the prediction task.

\section{Discussion and Conclusion}

Together, our experiments establish a normative bias in aligned language models: alignment improves prediction where human behavior tracks normative theory (one-shot games, non-strategic lotteries), and degrades it where behavior diverges from norms (multi-round strategic interactions). RLHF and related methods optimize for responses that annotators approve of--cooperative, fair, socially appropriate--creating what might be called a normative model of behavior. Predicting actual strategic decisions, however, requires a descriptive model: one that captures choices people may not endorse, including bluffing, retaliation, and history-dependent deviations from cooperation.

The selective nature of the aligned-model advantage rules out the most natural alternative explanation: if alignment merely degraded general capabilities (catastrophic forgetting), aligned models would underperform uniformly rather than winning selectively on one-shot games and lotteries. The relevant knowledge is preserved; alignment shifts \emph{which} behavioral patterns the model expresses, not whether it can express them at all.

The distributional narrowing documented in Section~2.2 offers a precise account. KL-regularized reward maximization yields an optimal policy $\pi^*(x) \propto \pi_0(x)\exp(r(x)/\beta)$--an exponential tilt of the base distribution that concentrates mass on high-reward (annotator-approved) behavioral modes at the expense of the tails \citep{korbak2022rl}. \citet{xiao2024bias} showed that this concentration is not a side effect but a structural property: standard RLHF exhibits an inherent bias toward dominant preferences (``preference collapse''), and preserving the full preference distribution would require an entropy-based regularizer that current methods lack. Our results provide the first behavioral evidence for this theoretical prediction--the collapse is not merely measurable in generation diversity \citep{kirk2024understanding} but in predictive fidelity for human decisions. The tails that reward tilting suppresses are precisely where multi-round strategic behavior lives: reciprocity, retaliation, and reputation dynamics that annotators would not endorse but that humans routinely exhibit.

These findings carry practical implications in both directions. For multi-round interactive settings, base models should be preferred; for one-shot games or non-strategic tasks, aligned models remain appropriate. More broadly, alignment systematically narrows the behavioral distribution that pre-trained models encode, and any application that relies on LLMs to represent how people \emph{actually} behave faces the same risk. Researchers simulating voter behavior, consumer choices, or social media dynamics with aligned models may obtain results that reflect idealized rather than actual human behavior. The growing use of LLMs as simulated participants in social science \citep{horton2023homo, aher2023using} makes this an active methodological risk: studies reporting that ``LLMs replicate human behavior'' may in fact be reporting that LLMs replicate \emph{normative} behavior, with the gap invisible where norms and behavior coincide.

Several open questions follow naturally. Which aspects of multi-round play drive the base advantage--opponent modeling, history integration, or trajectory novelty? Extending to continuous negotiations, auctions, or coalition formation would test generality. From an alignment perspective, developing methods that preserve empirical behavioral distributions while adding helpfulness is a natural direction. Finally, testing whether the effect persists at extreme scale would clarify whether the normative shift is inherent to alignment or diminishes as models grow more capable.

The normative--descriptive trade-off documented here may be inherent to current alignment methods: optimizing for a single reward model that encodes annotator preferences cannot simultaneously preserve the full distribution of human behavior. Until alignment methods are developed that can add helpfulness without collapsing behavioral diversity, the choice of base versus aligned model is not merely a formatting decision but a substantive modeling assumption--one that determines whether an LLM serves as a model \emph{of} human behavior or a model \emph{for} human use.

\section*{Limitations}

First, the GLEE multi-round game data comes from human participants playing against LLM opponents \citep{shapira2024glee}, not other humans. However, participants were not informed that their opponent was an LLM (GLEE presented the other player by name), and matrix game participants \citep{akata2025playing} were told they might face either a human or an artificial agent--so human decisions were made without certain knowledge of the opponent's nature, mitigating concerns about altered behavior. Second, our analysis is restricted to binary or ternary decisions; whether the findings extend to continuous action spaces remains open. Third, all 120 pairs are open-weight; we cannot evaluate closed-source models for which base versions are unavailable, though consistent trends from 1B to 70B+ suggest the effect may generalize. Fourth, the one-shot boundary condition uses a different dataset \citep{zhu2025predicting} than the repeated games; the round-1 aligned-model advantage within multi-round games provides convergent evidence from the same data. Finally, we cannot rule out all alternative mechanisms, though the aligned-model advantage on one-shot games (Section~\ref{sec:boundary}) argues against catastrophic forgetting as the primary explanation.

\section*{Acknowledgments}

Eilam Shapira is supported by a Google PhD Fellowship. Roi Reichart has been partially supported by a VATAT grant on data science. We thank Maya Zadok, Alan Arazi, and Nitay Calderon for valuable feedback on earlier drafts.

\bibliography{references}

@inproceedings{horton2023homo,
  author    = {Filippas, Apostolos and Horton, John J. and Manning, Benjamin S.},
  title     = {Large Language Models as Simulated Economic Agents: What Can We Learn from {Homo Silicus}?},
  booktitle = {Proceedings of the 25th {ACM} Conference on Economics and Computation, {EC} 2024},
  pages     = {614--615},
  year      = {2024},
  publisher = {{ACM}},
  doi       = {10.1145/3670865.3673513},
}

@inproceedings{aher2023using,
  author    = {Aher, Gati and Arriaga, Rosa I. and Kalai, Adam Tauman},
  title     = {Using large language models to simulate multiple humans and replicate human subject studies},
  year      = {2023},
  publisher = {JMLR.org},
  booktitle = {Proceedings of the 40th International Conference on Machine Learning},
  articleno = {17},
  numpages  = {35},
  location  = {Honolulu, Hawaii, USA},
  series    = {ICML'23},
}

@article{argyle2023out,
  author    = {Argyle, Lisa P. and Busby, Ethan C. and Fulda, Nancy and Gubler, Joshua R. and Rytting, Christopher and Wingate, David},
  title     = {Out of One, Many: Using Language Models to Simulate Human Samples},
  journal   = {Political Analysis},
  volume    = {31},
  number    = {3},
  pages     = {337--351},
  year      = {2023},
  publisher = {Cambridge University Press},
  doi       = {10.1017/pan.2023.2},
}

@article{binz2023using,
  author    = {Binz, Marcel and Schulz, Eric},
  title     = {Using Cognitive Psychology to Understand {GPT-3}},
  journal   = {Proceedings of the National Academy of Sciences},
  volume    = {120},
  number    = {6},
  pages     = {e2218523120},
  year      = {2023},
  doi       = {10.1073/pnas.2218523120},
}

@unpublished{hewitt2024predicting,
  author    = {Hewitt, Luke and Ashokkumar, Ashwini and Ghezae, Isaias and Willer, Robb},
  title     = {Predicting Results of Social Science Experiments Using Large Language Models},
  year      = {2024},
  note      = {Working paper},
}

@inproceedings{suh2025subpop,
  author    = {Suh, Joseph and Jahanparast, Erfan and Moon, Suhong and Kang, Minwoo and Chang, Serina},
  title     = {Language Model Fine-Tuning on Scaled Survey Data for Predicting Distributions of Public Opinions},
  booktitle = {Proceedings of the 63rd Annual Meeting of the Association for Computational Linguistics (Volume 1: Long Papers)},
  year      = {2025},
  publisher = {Association for Computational Linguistics},
  pages     = {21147--21170},
  doi       = {10.18653/v1/2025.acl-long.1028},
}

@inproceedings{ouyang2022training,
  author    = {Ouyang, Long and Wu, Jeffrey and Jiang, Xu and Almeida, Diogo and Wainwright, Carroll L. and Mishkin, Pamela and Zhang, Chong and Agarwal, Sandhini and Slama, Katarina and Ray, Alex and Schulman, John and Hilton, Jacob and Kelton, Fraser and Miller, Luke and Simens, Maddie and Askell, Amanda and Welinder, Peter and Christiano, Paul F. and Leike, Jan and Lowe, Ryan},
  title     = {Training Language Models to Follow Instructions with Human Feedback},
  booktitle = {Advances in Neural Information Processing Systems},
  volume    = {35},
  pages     = {27730--27744},
  year      = {2022},
}

@inproceedings{rafailov2023direct,
  author    = {Rafailov, Rafael and Sharma, Archit and Mitchell, Eric and Manning, Christopher D. and Ermon, Stefano and Finn, Chelsea},
  title     = {Direct Preference Optimization: Your Language Model is Secretly a Reward Model},
  booktitle = {Advances in Neural Information Processing Systems},
  volume    = {36},
  pages     = {53728--53741},
  year      = {2023},
}

@article{munjal2026base,
  author    = {Munjal, Prateek and Christophe, Clement and Rajan, Ronnie and Kanithi, Praveenkumar},
  title     = {Do Instruction-Tuned Models Always Perform Better Than Base Models? {E}vidence from Math and Domain-Shifted Benchmarks},
  journal   = {arXiv preprint arXiv:2601.13244},
  year      = {2026},
}

@article{kadavath2022language,
  author    = {Kadavath, Saurav and Conerly, Tom and Askell, Amanda and Henighan, Tom and Drain, Dawn and Perez, Ethan and Schiefer, Nicholas and Hatfield-Dodds, Zac and DasSarma, Nova and Tran-Johnson, Eli and Johnston, Scott and El-Showk, Sheer and Jones, Andy and Elhage, Nelson and Hume, Tristan and Chen, Anna and Bai, Yuntao and Bowman, Sam and Fort, Stanislav and Ganguli, Deep and Hernandez, Danny and Jacobson, Josh and Kernion, Jackson and Kravec, Shauna and Lovitt, Liane and Ndousse, Kamal and Olsson, Catherine and Ringer, Sam and Amodei, Dario and Brown, Tom and Clark, Jack and Joseph, Nicholas and Mann, Ben and McCandlish, Sam and Olah, Chris and Kaplan, Jared},
  title     = {Language Models (Mostly) Know What They Know},
  journal   = {arXiv preprint arXiv:2207.05221},
  year      = {2022},
}

@inproceedings{zhu2023calibration,
  author    = {Zhu, Chiwei and Xu, Benfeng and Wang, Quan and Zhang, Yongdong and Mao, Zhendong},
  title     = {On the Calibration of Large Language Models and Alignment},
  booktitle = {Findings of the Association for Computational Linguistics: EMNLP 2023},
  year      = {2023},
  publisher = {Association for Computational Linguistics},
  pages     = {9778--9795},
  doi       = {10.18653/v1/2023.findings-emnlp.654},
}

@inproceedings{kirk2024understanding,
  author    = {Kirk, Robert and Mediratta, Ishita and Nalmpantis, Christoforos and Luketina, Jelena and Hambro, Eric and Grefenstette, Edward and Raileanu, Roberta},
  title     = {Understanding the Effects of {RLHF} on {LLM} Generalisation and Diversity},
  booktitle = {The Twelfth International Conference on Learning Representations},
  year      = {2024},
  url       = {https://openreview.net/forum?id=PXD3FAVHJT},
}

@inproceedings{cao2025entropy,
  author    = {Cao, Steven and Valiant, Gregory and Liang, Percy},
  title     = {On the Entropy Calibration of Language Models},
  booktitle = {Advances in Neural Information Processing Systems 38},
  year      = {2025},
  url       = {https://openreview.net/forum?id=ZpQ2SqQNXf},
}

@inproceedings{chen2025kl,
  author    = {Anthony GX-Chen and Jatin Prakash and Jeff Guo and Rob Fergus and Rajesh Ranganath},
  title     = {{KL}-Regularized Reinforcement Learning is Designed to Mode Collapse},
  booktitle = {The Fourteenth International Conference on Learning Representations},
  year      = {2026},
  url       = {https://openreview.net/forum?id=flBRtdIihA},
}

@inproceedings{santurkar2023whose,
  author    = {Santurkar, Shibani and Durmus, Esin and Ladhak, Faisal and Lee, Cinoo and Liang, Percy and Hashimoto, Tatsunori},
  title     = {Whose Opinions Do Language Models Reflect?},
  booktitle = {Proceedings of the 40th International Conference on Machine Learning},
  series    = {Proceedings of Machine Learning Research},
  volume    = {202},
  pages     = {29971--30004},
  year      = {2023},
  publisher = {PMLR},
}

@article{itzhak2024instructed,
  author    = {Itzhak, Itay and Stanovsky, Gabriel and Rosenfeld, Nir and Belinkov, Yonatan},
  title     = {Instructed to Bias: Instruction-Tuned Language Models Exhibit Emergent Cognitive Bias},
  journal   = {Transactions of the Association for Computational Linguistics},
  volume    = {12},
  pages     = {771--785},
  year      = {2024},
  publisher = {MIT Press},
  doi       = {10.1162/tacl_a_00673},
}

@inproceedings{liu2025llms,
  author    = {Liu, Ryan and Geng, Jiayi and Peterson, Joshua C. and Sucholutsky, Ilia and Griffiths, Thomas L.},
  title     = {Large Language Models Assume People are More Rational than We Really are},
  booktitle = {Proceedings of the Thirteenth International Conference on Learning Representations},
  year      = {2025},
  url       = {https://openreview.net/forum?id=dAeET8gxqg},
}

@article{tjuatja2024llms,
  author    = {Tjuatja, Lindia and Chen, Valerie and Wu, Tongshuang and Talwalkar, Ameet and Neubig, Graham},
  title     = {Do {LLMs} Exhibit Human-like Response Biases? {A} Case Study in Survey Design},
  journal   = {Transactions of the Association for Computational Linguistics},
  volume    = {12},
  pages     = {1011--1026},
  year      = {2024},
  publisher = {MIT Press},
  doi       = {10.1162/tacl_a_00685},
}

@article{capraro2025optimistic,
  author    = {Capraro, Valerio and Di Paolo, Roberto and Pizziol, Veronica},
  title     = {A Publicly Available Benchmark for Assessing Large Language Models' Ability to Predict How Humans Balance Self-Interest and the Interest of Others},
  journal   = {Scientific Reports},
  volume    = {15},
  pages     = {21428},
  year      = {2025},
  publisher = {Nature Publishing Group},
  doi       = {10.1038/s41598-025-01715-7},
}

@inproceedings{bauer2025gpt4,
  author    = {Bauer, Kevin and Liebich, Lena and Kosfeld, Michael},
  title     = {Can {GPT} Mimic Human Preferences? {An} Empirical and Structural Investigation},
  booktitle = {Proceedings of the 33rd European Conference on Information Systems (ECIS 2025)},
  year      = {2025},
  url       = {https://aisel.aisnet.org/ecis2025/ai_anthro/ai_anthro/4/},
}

@article{shapira2026poisoned,
  author    = {Shapira, Eilam and Tennenholtz, Moshe and Reichart, Roi},
  title     = {The Poisoned Apple Effect: Strategic Manipulation of Mediated Markets via Technology Expansion of {AI} Agents},
  journal   = {arXiv preprint arXiv:2601.11496},
  year      = {2026},
}

@article{mei2024turing,
  author    = {Mei, Qiaozhu and Xie, Yutong and Yuan, Walter and Jackson, Matthew O.},
  title     = {A {Turing} Test of Whether {AI} Chatbots Are Behaviorally Similar to Humans},
  journal   = {Proceedings of the National Academy of Sciences},
  volume    = {121},
  number    = {9},
  pages     = {e2313925121},
  year      = {2024},
  doi       = {10.1073/pnas.2313925121},
}

@article{akata2025playing,
  author    = {Akata, Elif and Schulz, Lion and Coda-Forno, Julian and Oh, Seong Joon and Bethge, Matthias and Schulz, Eric},
  title     = {Playing Repeated Games with Large Language Models},
  journal   = {Nature Human Behaviour},
  volume    = {9},
  number    = {7},
  pages     = {1380--1390},
  year      = {2025},
  publisher = {Nature Publishing Group},
  doi       = {10.1038/s41562-025-02172-y},
}

@article{mckelvey1995quantal,
  author    = {McKelvey, Richard D. and Palfrey, Thomas R.},
  title     = {Quantal Response Equilibria for Normal Form Games},
  journal   = {Games and Economic Behavior},
  volume    = {10},
  number    = {1},
  pages     = {6--38},
  year      = {1995},
  publisher = {Elsevier},
}

@article{mckelvey1998quantal,
  author    = {McKelvey, Richard D. and Palfrey, Thomas R.},
  title     = {Quantal Response Equilibria for Extensive Form Games},
  journal   = {Experimental Economics},
  volume    = {1},
  number    = {1},
  pages     = {9--41},
  year      = {1998},
  publisher = {Springer},
}

@article{nagel1995unraveling,
  author    = {Nagel, Rosemarie},
  title     = {Unraveling in Guessing Games: An Experimental Study},
  journal   = {American Economic Review},
  volume    = {85},
  number    = {5},
  pages     = {1313--1326},
  year      = {1995},
}

@article{stahl1995players,
  author    = {Stahl, Dale O. and Wilson, Paul W.},
  title     = {On Players' Models of Other Players: Theory and Experimental Evidence},
  journal   = {Games and Economic Behavior},
  volume    = {10},
  number    = {1},
  pages     = {218--254},
  year      = {1995},
  publisher = {Elsevier},
}

@article{camerer2004cognitive,
  author    = {Camerer, Colin F. and Ho, Teck-Hua and Chong, Juin-Kuan},
  title     = {A Cognitive Hierarchy Model of Games},
  journal   = {The Quarterly Journal of Economics},
  volume    = {119},
  number    = {3},
  pages     = {861--898},
  year      = {2004},
  publisher = {MIT Press},
}

@article{camerer1999experience,
  author    = {Camerer, Colin F. and Ho, Teck-Hua},
  title     = {Experience-Weighted Attraction Learning in Normal Form Games},
  journal   = {Econometrica},
  volume    = {67},
  number    = {4},
  pages     = {827--874},
  year      = {1999},
}

@article{zhu2025predicting,
  author    = {Zhu, Jian-Qiao and Peterson, Joshua C. and Enke, Benjamin and Griffiths, Thomas L.},
  title     = {Capturing the Complexity of Human Strategic Decision-Making with Machine Learning},
  journal   = {Nature Human Behaviour},
  volume    = {9},
  pages     = {2114--2120},
  year      = {2025},
  publisher = {Nature Publishing Group},
  doi       = {10.1038/s41562-025-02230-5},
}

@article{shapira2024llm,
  author    = {Shapira, Eilam and Madmon, Omer and Reichart, Roi and Tennenholtz, Moshe},
  title     = {Can {LLMs} Replace Economic Choice Prediction Labs? {T}he Case of Language-Based Persuasion Games},
  journal   = {arXiv preprint arXiv:2401.17435},
  year      = {2024},
}

@article{shapira2024glee,
  author    = {Shapira, Eilam and Madmon, Omer and Reinman, Itamar and Amouyal, Samuel Joseph and Reichart, Roi and Tennenholtz, Moshe},
  title     = {{GLEE}: A Unified Framework and Benchmark for Language-based Economic Environments},
  journal   = {arXiv preprint arXiv:2410.05254},
  year      = {2024},
}

@article{shapira2025prediction,
  author    = {Shapira, Eilam and Madmon, Omer and Apel, Reut and Tennenholtz, Moshe and Reichart, Roi},
  title     = {Human Choice Prediction in Language-based Persuasion Games: Simulation-based Off-Policy Evaluation},
  journal   = {Transactions of the Association for Computational Linguistics},
  volume    = {13},
  pages     = {980--1006},
  year      = {2025},
  publisher = {MIT Press},
}

@article{rubinstein1982perfect,
  author    = {Rubinstein, Ariel},
  title     = {Perfect Equilibrium in a Bargaining Model},
  journal   = {Econometrica},
  volume    = {50},
  number    = {1},
  pages     = {97--109},
  year      = {1982},
}

@article{crawford1982strategic,
  author    = {Crawford, Vincent P. and Sobel, Joel},
  title     = {Strategic Information Transmission},
  journal   = {Econometrica},
  volume    = {50},
  number    = {6},
  pages     = {1431--1451},
  year      = {1982},
}

@inproceedings{korbak2022rl,
  author    = {Korbak, Tomasz and Perez, Ethan and Buckley, Christopher},
  title     = {{RL} with {KL} Penalties is Better Viewed as {B}ayesian Inference},
  booktitle = {Findings of the Association for Computational Linguistics: EMNLP 2022},
  year      = {2022},
  publisher = {Association for Computational Linguistics},
  pages     = {1083--1091},
  doi       = {10.18653/v1/2022.findings-emnlp.77},
}

@article{xiao2024bias,
  author    = {Xiao, Jiancong and Li, Ziniu and Xie, Xingyu and Getzen, Emily and Fang, Cong and Long, Qi and Su, Weijie J.},
  title     = {On the Algorithmic Bias of Aligning Large Language Models with {RLHF}: Preference Collapse and Matching Regularization},
  journal   = {Journal of the American Statistical Association},
  volume    = {120},
  number    = {552},
  pages     = {2154--2164},
  year      = {2025},
  doi       = {10.1080/01621459.2025.2555067},
}

@inproceedings{marantz2025binary,
  author    = {Marantz, Eyal and Plonsky, Ori},
  title     = {Predicting Human Choice Between Textually Described Lotteries},
  booktitle = {Proceedings of the 47th Annual Conference of the Cognitive Science Society},
  year      = {2025},
  note      = {arXiv:2503.14004},
}

\clearpage
\appendix

\section{Full Model Inventory}
\label{app:models}

This appendix supplements the model description in Section~\ref{sec:intro}. Table~\ref{tab:models} lists all 120 same-provider base--aligned model pairs used in our experiments, grouped by family and sorted by parameter count. Same-provider means both models are released by the same organization on HuggingFace. We exclude pairs where the base and aligned checkpoints are identical, and pairs where the aligned model lacks a chat template (required for the native-format comparison).

\section{Prompt Content Variants}
\label{app:variants}

Section~\ref{sec:prompt_robustness} examines 14 prompt variants organized into four clusters. Table~\ref{tab:variants} details each variant.

The four failed variants demonstrate that base models critically depend on structured completion suffixes to concentrate probability mass on decision tokens. The \emph{simplified} and \emph{minimal} variants, which replace the JSON pattern with unstructured suffixes, cause decision token mass to drop from $\sim$90\% to below 10\%. The \emph{natural language} variant retains marginally higher mass but too few pairs survive standard filtering (mass $\geq 0.8$). The \emph{numbers only} variant, which strips dialogue history, similarly yields too few valid pairs ($N \leq 5$).

\section{Filtering Criteria and Sensitivity}
\label{app:filter_details}
\label{app:sensitivity}

This appendix details the two pair-level filters summarized in Section~\ref{sec:filtering} and demonstrates that the base advantage is robust to the choice of filtering thresholds.

\paragraph{Mass filter.} For each model and game family, we compute the average probability mass on decision tokens across all decision points--the sum of softmax probabilities assigned to recognized decision tokens (e.g., ``accept'' and ``reject''). If either model in a pair falls below an average mass of 0.8 on a given family, both models are excluded from that family. Models below this threshold do not reliably produce decision-relevant tokens, making their normalized probabilities unreliable.

\paragraph{Minimum correlation filter.} For each model pair and game family, we compute the Pearson correlation between each model's $p_{\text{accept}}$ predictions and the actual binary human decisions. If \emph{both} models in the pair fall below a Pearson correlation of 0.3, the pair is excluded from that family. If at least one model exceeds the threshold, both are retained. This removes uninformative pairs where neither model predicts above a minimal threshold, ensuring that base-vs-aligned comparisons reflect genuine differences in predictive quality rather than noise from two equally poor models.

Both filters are applied independently per game family, so a pair excluded from bargaining may still contribute to persuasion or negotiation analyses.

\paragraph{Sensitivity analysis.} Tables~\ref{tab:sensitivity_bargaining}--\ref{tab:sensitivity_overall} show that the base advantage is robust across all four game families and a wide range of mass and correlation threshold choices: in every cell of every table, base models win the majority of comparisons with $p < 0.05$. The cell corresponding to our chosen thresholds (mass $\geq 0.8$, correlation $\geq 0.3$) is highlighted in bold.

\section{Per-Pair Prediction Results}
\label{app:per_pair}

This appendix supplements Section~\ref{sec:results}. Tables~\ref{tab:perpair_bpn}--\ref{tab:perpair_matrix_boundary} list the average decision-token mass and Pearson correlation with human decisions for every same-provider pair across all six datasets: the four main game families (with PD and BoS combined into a single matrix vector per pair) and the two boundary condition datasets. Pairs are numbered consistently across tables and correspond to the model inventory in Appendix~\ref{app:models}.

\section{Game Configuration Robustness}
\label{app:configs}

This appendix supplements the game configuration robustness analysis in Section~\ref{sec:results}. Each GLEE game family is parameterized along multiple dimensions. Table~\ref{tab:params} documents the full parameter space. Tables~\ref{tab:config_barg}--\ref{tab:config_matrix} present base-vs-aligned win counts for every parameter value in each game family. $N$ = valid pairs after filtering; Filt.\ = pairs excluded by mass or correlation filters. The base advantage is consistent across all parameter values and all families. The sole exception is bargaining with discount factor $\delta_1 = 0.8$ (the most impatient proposer), where the advantage narrows to near parity (10:7, $p = 0.31$).

\onecolumn
\section*{Appendix Tables}

{\footnotesize
\begin{longtable}{@{}rllr@{}}
\caption{All base--aligned pairs, grouped alphabetically by family and sorted by parameter count within each family.}\label{tab:models}\\
\toprule
\textbf{\#} & \textbf{Base Model} & \textbf{Aligned Model} & \textbf{Size} \\
\midrule
\endfirsthead
\toprule
\textbf{\#} & \textbf{Base Model} & \textbf{Aligned Model} & \textbf{Size} \\
\midrule
\endhead
\midrule
\multicolumn{4}{r}{\textit{Continued on next page}} \\
\endfoot
\bottomrule
\endlastfoot
\multicolumn{4}{l}{\textbf{CodeGemma} (1 pair)} \\
1 & codegemma-7b & codegemma-7b-it & 7B \\
\multicolumn{4}{l}{\textbf{CodeLlama} (6 pairs)} \\
2 & CodeLlama-7b-hf & CodeLlama-7b-Instruct-hf & 7B \\
3 & CodeLlama-7b-Python-hf & CodeLlama-7b-Instruct-hf & 7B \\
4 & CodeLlama-13b-hf & CodeLlama-13b-Instruct-hf & 13B \\
5 & CodeLlama-13b-Python-hf & CodeLlama-13b-Instruct-hf & 13B \\
6 & CodeLlama-34b-hf & CodeLlama-34b-Instruct-hf & 34B \\
7 & CodeLlama-34b-Python-hf & CodeLlama-34b-Instruct-hf & 34B \\
\multicolumn{4}{l}{\textbf{DeepSeek} (6 pairs)} \\
8 & deepseek-coder-1.3b-base & deepseek-coder-1.3b-instruct & 1.3B \\
9 & deepseek-coder-6.7b-base & deepseek-coder-6.7b-instruct & 6.7B \\
10 & deepseek-llm-7b-base & deepseek-llm-7b-chat & 7B \\
11 & deepseek-math-7b-base & deepseek-math-7b-instruct & 7B \\
12 & deepseek-coder-33b-base & deepseek-coder-33b-instruct & 33B \\
13 & deepseek-llm-67b-base & deepseek-llm-67b-chat & 67B \\
\multicolumn{4}{l}{\textbf{Falcon} (11 pairs)} \\
14 & Falcon-H1-0.5B-Base & Falcon-H1-0.5B-Instruct & 0.5B \\
15 & Falcon3-1B-Base & Falcon3-1B-Instruct & 1B \\
16 & Falcon-H1-1.5B-Base & Falcon-H1-1.5B-Instruct & 1.5B \\
17 & Falcon3-3B-Base & Falcon3-3B-Instruct & 3B \\
18 & Falcon-H1-3B-Base & Falcon-H1-3B-Instruct & 3B \\
19 & falcon-mamba-7b & falcon-mamba-7b-instruct & 7B \\
20 & Falcon3-7B-Base & Falcon3-7B-Instruct & 7B \\
21 & Falcon3-Mamba-7B-Base & Falcon3-Mamba-7B-Instruct & 7B \\
22 & Falcon-H1-7B-Base & Falcon-H1-7B-Instruct & 7B \\
23 & Falcon3-10B-Base & Falcon3-10B-Instruct & 10B \\
24 & Falcon-H1-34B-Base & Falcon-H1-34B-Instruct & 34B \\
\multicolumn{4}{l}{\textbf{Gemma} (11 pairs)} \\
25 & gemma-3-1b-pt & gemma-3-1b-it & 1B \\
26 & gemma-2-2b & gemma-2-2b-it & 2B \\
27 & gemma-2b & gemma-2b-it & 2B \\
28 & recurrentgemma-2b & recurrentgemma-2b-it & 2B \\
29 & gemma-3-4b-pt & gemma-3-4b-it & 4B \\
30 & gemma-7b & gemma-7b-it & 7B \\
31 & gemma-2-9b & gemma-2-9b-it & 9B \\
32 & recurrentgemma-9b & recurrentgemma-9b-it & 9B \\
33 & gemma-3-12b-pt & gemma-3-12b-it & 12B \\
34 & gemma-2-27b & gemma-2-27b-it & 27B \\
35 & gemma-3-27b-pt & gemma-3-27b-it & 27B \\
\multicolumn{4}{l}{\textbf{Granite} (5 pairs)} \\
36 & granite-3.0-2b-base & granite-3.0-2b-instruct & 2B \\
37 & granite-3.1-2b-base & granite-3.1-2b-instruct & 2B \\
38 & granite-3.0-8b-base & granite-3.0-8b-instruct & 8B \\
39 & granite-3.1-8b-base & granite-3.1-8b-instruct & 8B \\
40 & granite-20b-code-base & granite-20b-code-instruct & 20B \\
\multicolumn{4}{l}{\textbf{H2O} (3 pairs)} \\
41 & h2o-danube3-500m-base & h2o-danube3-500m-chat & 0.5B \\
42 & h2o-danube2-1.8b-base & h2o-danube2-1.8b-chat & 1.8B \\
43 & h2o-danube3-4b-base & h2o-danube3-4b-chat & 4B \\
\multicolumn{4}{l}{\textbf{Llama} (8 pairs)} \\
44 & Llama-3.2-1B & Llama-3.2-1B-Instruct & 1B \\
45 & Llama-3.2-3B & Llama-3.2-3B-Instruct & 3B \\
46 & Llama-2-7b-hf & Llama-2-7b-chat-hf & 7B \\
47 & Meta-Llama-3.1-8B & Meta-Llama-3.1-8B-Instruct & 8B \\
48 & Meta-Llama-3-8B & Meta-Llama-3-8B-Instruct & 8B \\
49 & Llama-2-13b-hf & Llama-2-13b-chat-hf & 13B \\
50 & Meta-Llama-3-70B & Meta-Llama-3-70B-Instruct & 70B \\
51 & Meta-Llama-3.1-70B & Meta-Llama-3.1-70B-Instruct & 70B \\
\multicolumn{4}{l}{\textbf{MAP-Neo} (1 pair)} \\
52 & neo\_7b & neo\_7b\_instruct\_v0.1 & 7B \\
\multicolumn{4}{l}{\textbf{MiMo} (1 pair)} \\
53 & MiMo-7B-Base & MiMo-7B-RL & 7B \\
\multicolumn{4}{l}{\textbf{Mistral} (5 pairs)} \\
54 & Mistral-7B-v0.3 & Mistral-7B-Instruct-v0.3 & 7B \\
55 & Mistral-7B-v0.1 & Mistral-7B-Instruct-v0.2 & 7B \\
56 & Mistral-7B-v0.1 & Mistral-7B-Instruct-v0.1 & 7B \\
57 & Mistral-Nemo-Base-2407 & Mistral-Nemo-Instruct-2407 & 12B \\
58 & Mistral-Small-24B-Base-2501 & Mistral-Small-24B-Instruct-2501 & 24B \\
\multicolumn{4}{l}{\textbf{Nemotron} (1 pair)} \\
59 & Minitron-4B-Base & Nemotron-Mini-4B-Instruct & 4B \\
\multicolumn{4}{l}{\textbf{OLMo} (8 pairs)} \\
60 & OLMo-2-0425-1B & OLMo-2-0425-1B-Instruct & 1B \\
61 & OLMo-7B-hf & OLMo-7B-Instruct-hf & 7B \\
62 & OLMo-2-1124-7B & OLMo-2-1124-7B-Instruct & 7B \\
63 & Olmo-3-1025-7B & Olmo-3-7B-Instruct & 7B \\
64 & OLMoE-1B-7B-0125 & OLMoE-1B-7B-0125-Instruct & 7B \\
65 & OLMo-2-1124-13B & OLMo-2-1124-13B-Instruct & 13B \\
66 & OLMo-2-0325-32B & OLMo-2-0325-32B-Instruct & 32B \\
67 & Olmo-3-1125-32B & Olmo-3.1-32B-Instruct & 32B \\
\multicolumn{4}{l}{\textbf{Qwen} (32 pairs)} \\
68 & Qwen2.5-0.5B & Qwen2.5-0.5B-Instruct & 0.5B \\
69 & Qwen2-0.5B & Qwen2-0.5B-Instruct & 0.5B \\
70 & Qwen1.5-0.5B & Qwen1.5-0.5B-Chat & 0.5B \\
71 & Qwen2.5-Coder-0.5B & Qwen2.5-Coder-0.5B-Instruct & 0.5B \\
72 & Qwen3-0.6B-Base & Qwen3-0.6B & 0.6B \\
73 & Qwen2.5-1.5B & Qwen2.5-1.5B-Instruct & 1.5B \\
74 & Qwen2-1.5B & Qwen2-1.5B-Instruct & 1.5B \\
75 & Qwen2.5-Math-1.5B & Qwen2.5-Math-1.5B-Instruct & 1.5B \\
76 & Qwen2.5-Coder-1.5B & Qwen2.5-Coder-1.5B-Instruct & 1.5B \\
77 & Qwen3-1.7B-Base & Qwen3-1.7B & 1.7B \\
78 & Qwen1.5-1.8B & Qwen1.5-1.8B-Chat & 1.8B \\
79 & Qwen2.5-3B & Qwen2.5-3B-Instruct & 3B \\
80 & Qwen2.5-Coder-3B & Qwen2.5-Coder-3B-Instruct & 3B \\
81 & Qwen1.5-4B & Qwen1.5-4B-Chat & 4B \\
82 & Qwen3-4B-Base & Qwen3-4B & 4B \\
83 & Qwen3-4B-Base & Qwen3-4B-Instruct-2507 & 4B \\
84 & Qwen2.5-7B & Qwen2.5-7B-Instruct & 7B \\
85 & Qwen2-7B & Qwen2-7B-Instruct & 7B \\
86 & Qwen1.5-7B & Qwen1.5-7B-Chat & 7B \\
87 & Qwen2.5-Math-7B & Qwen2.5-Math-7B-Instruct & 7B \\
88 & Qwen2.5-Coder-7B & Qwen2.5-Coder-7B-Instruct & 7B \\
89 & Qwen3-8B-Base & Qwen3-8B & 8B \\
90 & Qwen2.5-14B & Qwen2.5-14B-Instruct & 14B \\
91 & Qwen1.5-14B & Qwen1.5-14B-Chat & 14B \\
92 & Qwen1.5-MoE-A2.7B & Qwen1.5-MoE-A2.7B-Chat & 14B \\
93 & Qwen3-14B-Base & Qwen3-14B & 14B \\
94 & Qwen2.5-Coder-14B & Qwen2.5-Coder-14B-Instruct & 14B \\
95 & Qwen2.5-32B & Qwen2.5-32B-Instruct & 32B \\
96 & Qwen1.5-32B & Qwen1.5-32B-Chat & 32B \\
97 & Qwen2.5-Coder-32B & Qwen2.5-Coder-32B-Instruct & 32B \\
98 & Qwen2-72B & Qwen2-72B-Instruct & 72B \\
99 & Qwen2.5-72B & Qwen2.5-72B-Instruct & 72B \\
\multicolumn{4}{l}{\textbf{Sailor} (2 pairs)} \\
100 & Sailor-4B & Sailor-4B-Chat & 4B \\
101 & Sailor-7B & Sailor-7B-Chat & 7B \\
\multicolumn{4}{l}{\textbf{SeaLLM} (1 pair)} \\
102 & SeaLLM-7B-v2 & SeaLLM-7B-v2.5 & 7B \\
\multicolumn{4}{l}{\textbf{Seed-Coder} (1 pair)} \\
103 & Seed-Coder-8B-Base & Seed-Coder-8B-Instruct & 8B \\
\multicolumn{4}{l}{\textbf{SmolLM} (6 pairs)} \\
104 & SmolLM-135M & SmolLM-135M-Instruct & 0.1B \\
105 & SmolLM2-135M & SmolLM2-135M-Instruct & 0.1B \\
106 & SmolLM-360M & SmolLM-360M-Instruct & 0.4B \\
107 & SmolLM2-360M & SmolLM2-360M-Instruct & 0.4B \\
108 & SmolLM-1.7B & SmolLM-1.7B-Instruct & 1.7B \\
109 & SmolLM2-1.7B & SmolLM2-1.7B-Instruct & 1.7B \\
\multicolumn{4}{l}{\textbf{Solar} (1 pair)} \\
110 & SOLAR-10.7B-v1.0 & SOLAR-10.7B-Instruct-v1.0 & 10.7B \\
\multicolumn{4}{l}{\textbf{StableLM} (3 pairs)} \\
111 & stablelm-2-1\_6b & stablelm-2-1\_6b-chat & 1.6B \\
112 & stablelm-3b-4e1t & stablelm-zephyr-3b & 3B \\
113 & stablelm-2-12b & stablelm-2-12b-chat & 12B \\
\multicolumn{4}{l}{\textbf{TinyLlama} (1 pair)} \\
114 & TinyLlama-1.1B-intermediate-step-1431k-3T & TinyLlama-1.1B-Chat-v1.0 & 1.1B \\
\multicolumn{4}{l}{\textbf{Yi} (5 pairs)} \\
115 & Yi-1.5-6B & Yi-1.5-6B-Chat & 6B \\
116 & Yi-6B & Yi-6B-Chat & 6B \\
117 & Yi-1.5-9B & Yi-1.5-9B-Chat & 9B \\
118 & Yi-34B & Yi-34B-Chat & 34B \\
119 & Yi-1.5-34B & Yi-1.5-34B-Chat & 34B \\
\multicolumn{4}{l}{\textbf{Zamba2} (1 pair)} \\
120 & Zamba2-1.2B & Zamba2-1.2B-instruct & 1.2B \\
\end{longtable}
}

\begin{table}[H]
\centering
\footnotesize
\caption{All 14 prompt variants tested. Each modifies the reference JSON completion format. ``OK'' = sufficient valid pairs after filtering.}
\label{tab:variants}
\begin{tabular}{@{}llp{3.1cm}c@{}}
\toprule
\textbf{Cluster} & \textbf{Variant} & \textbf{Modification} & \textbf{OK} \\
\midrule
Baseline  & Standard          & Reference format & \checkmark \\
\midrule
Framing   & Predict human     & Sys: ``Predict what a participant decided'' & \checkmark \\
          & Observer          & Sys: external observer & \checkmark \\
          & Reversed roles    & Sys: offeror predicting receiver & \checkmark \\
\midrule
Persona   & Naive             & Sys: ``No prior experience'' & \checkmark \\
          & Expert            & Sys: ``Behavioral econ researcher'' & \checkmark \\
          & Fairness          & Sys: ``Values fairness'' & \checkmark \\
          & Selfish           & Sys: ``Maximize personal gain'' & \checkmark \\
          & Emotional         & Sys: ``Gut feeling'' & \checkmark \\
\midrule
Format    & Natural language  & Suffix: ``The decision is:~'' & \ding{55} \\
          & Simplified        & Suffix: ``Answer:~'' & \ding{55} \\
          & Minimal           & Suffix: ``I~'' & \ding{55} \\
\midrule
Structure & Numbers only      & Drops dialogue history & \ding{55} \\
          & Preamble rev.     & Swaps accept/reject order & \checkmark \\
\bottomrule
\end{tabular}
\end{table}


\begin{table}[H]
\centering
\caption{Sensitivity analysis: Bargaining -- Base vs.\ Aligned wins across mass and min-corr thresholds}
\label{tab:sensitivity_bargaining}
\resizebox{\columnwidth}{!}{%
\begin{tabular}{l c c c c c c}
\toprule
Mass $\downarrow$ / Min-Corr $\rightarrow$ & None & 0.1 & 0.2 & 0.3 & 0.4 & 0.5 \\
\midrule
None & 100:20 ($<10^{-14}$) & 96:10 ($<10^{-19}$) & 88:8 ($<10^{-18}$) & 79:6 ($<10^{-17}$) & 37:2 ($<10^{-9}$) & -- \\
0.5 & 93:16 ($<10^{-14}$) & 90:9 ($<10^{-18}$) & 83:8 ($<10^{-17}$) & 77:6 ($<10^{-17}$) & 37:2 ($<10^{-9}$) & -- \\
0.6 & 92:13 ($<10^{-16}$) & 90:8 ($<10^{-19}$) & 83:7 ($<10^{-18}$) & 77:5 ($<10^{-18}$) & 37:1 ($<10^{-10}$) & -- \\
0.7 & 91:9 ($<10^{-18}$) & 90:5 ($<10^{-21}$) & 83:5 ($<10^{-19}$) & 77:4 ($<10^{-19}$) & 37:1 ($<10^{-10}$) & -- \\
0.8 & 86:6 ($<10^{-19}$) & 86:5 ($<10^{-20}$) & 81:5 ($<10^{-19}$) & \textbf{75:4 ($<10^{-18}$)} & 36:1 ($<10^{-10}$) & -- \\
0.9 & 84:4 ($<10^{-21}$) & 84:4 ($<10^{-21}$) & 79:4 ($<10^{-19}$) & 73:4 ($<10^{-18}$) & 34:1 ($<10^{-9}$) & -- \\
\bottomrule
\end{tabular}
}
\end{table}

\begin{table}[H]
\centering
\caption{Sensitivity analysis: Persuasion -- Base vs.\ Aligned wins across mass and min-corr thresholds}
\label{tab:sensitivity_persuasion}
\resizebox{\columnwidth}{!}{%
\begin{tabular}{l c c c c c c}
\toprule
Mass $\downarrow$ / Min-Corr $\rightarrow$ & None & 0.1 & 0.2 & 0.3 & 0.4 & 0.5 \\
\midrule
None & 87:33 ($<10^{-7}$) & 80:32 ($<10^{-6}$) & 52:29 ($<10^{-3}$) & 46:7 ($<10^{-8}$) & -- & -- \\
0.5 & 61:31 ($<10^{-3}$) & 58:31 ($<10^{-3}$) & 40:28 ($0.09$) & 34:6 ($<10^{-6}$) & -- & -- \\
0.6 & 58:31 ($<10^{-3}$) & 56:31 ($<10^{-3}$) & 39:28 ($0.11$) & 33:6 ($<10^{-6}$) & -- & -- \\
0.7 & 44:22 ($<10^{-3}$) & 44:22 ($<10^{-3}$) & 38:21 ($0.02$) & 32:4 ($<10^{-7}$) & -- & -- \\
0.8 & 35:17 ($<10^{-3}$) & 35:17 ($<10^{-3}$) & 35:17 ($<10^{-3}$) & \textbf{32:4 ($<10^{-7}$)} & -- & -- \\
0.9 & 31:2 ($<10^{-8}$) & 31:2 ($<10^{-8}$) & 31:2 ($<10^{-8}$) & 28:1 ($<10^{-8}$) & -- & -- \\
\bottomrule
\end{tabular}
}
\end{table}

\begin{table}[H]
\centering
\caption{Sensitivity analysis: Negotiation -- Base vs.\ Aligned wins across mass and min-corr thresholds}
\label{tab:sensitivity_negotiation}
\resizebox{\columnwidth}{!}{%
\begin{tabular}{l c c c c c c}
\toprule
Mass $\downarrow$ / Min-Corr $\rightarrow$ & None & 0.1 & 0.2 & 0.3 & 0.4 & 0.5 \\
\midrule
None & 94:26 ($<10^{-10}$) & 76:5 ($<10^{-17}$) & 61:3 ($<10^{-15}$) & 35:1 ($<10^{-10}$) & 3:0 ($0.13$) & -- \\
0.5 & 83:16 ($<10^{-12}$) & 69:5 ($<10^{-16}$) & 55:3 ($<10^{-13}$) & 32:1 ($<10^{-9}$) & 3:0 ($0.13$) & -- \\
0.6 & 80:16 ($<10^{-11}$) & 67:5 ($<10^{-15}$) & 53:3 ($<10^{-13}$) & 31:1 ($<10^{-9}$) & 3:0 ($0.13$) & -- \\
0.7 & 77:14 ($<10^{-12}$) & 65:5 ($<10^{-14}$) & 52:3 ($<10^{-13}$) & 30:1 ($<10^{-8}$) & 3:0 ($0.13$) & -- \\
0.8 & 64:7 ($<10^{-13}$) & 56:2 ($<10^{-15}$) & 46:1 ($<10^{-13}$) & \textbf{25:1 ($<10^{-7}$)} & 3:0 ($0.13$) & -- \\
0.9 & 36:1 ($<10^{-10}$) & 33:1 ($<10^{-9}$) & 27:1 ($<10^{-7}$) & 16:1 ($<10^{-4}$) & 3:0 ($0.13$) & -- \\
\bottomrule
\end{tabular}
}
\end{table}

\begin{table}[H]
\centering
\caption{Sensitivity analysis: Matrix -- Base vs.\ Aligned wins across mass and min-corr thresholds}
\label{tab:sensitivity_matrix}
\resizebox{\columnwidth}{!}{%
\begin{tabular}{l c c c c c c}
\toprule
Mass $\downarrow$ / Min-Corr $\rightarrow$ & None & 0.1 & 0.2 & 0.3 & 0.4 & 0.5 \\
\midrule
None & 99:21 ($<10^{-13}$) & 98:20 ($<10^{-14}$) & 95:17 ($<10^{-14}$) & 81:13 ($<10^{-13}$) & 49:8 ($<10^{-8}$) & 4:0 ($0.06$) \\
0.5 & 99:20 ($<10^{-14}$) & 98:19 ($<10^{-14}$) & 95:17 ($<10^{-14}$) & 81:13 ($<10^{-13}$) & 49:8 ($<10^{-8}$) & 4:0 ($0.06$) \\
0.6 & 99:20 ($<10^{-14}$) & 98:19 ($<10^{-14}$) & 95:17 ($<10^{-14}$) & 81:13 ($<10^{-13}$) & 49:8 ($<10^{-8}$) & 4:0 ($0.06$) \\
0.7 & 99:20 ($<10^{-14}$) & 98:19 ($<10^{-14}$) & 95:17 ($<10^{-14}$) & 81:13 ($<10^{-13}$) & 49:8 ($<10^{-8}$) & 4:0 ($0.06$) \\
0.8 & 99:20 ($<10^{-14}$) & 98:19 ($<10^{-14}$) & 95:17 ($<10^{-14}$) & \textbf{81:13 ($<10^{-13}$)} & 49:8 ($<10^{-8}$) & 4:0 ($0.06$) \\
0.9 & 98:18 ($<10^{-15}$) & 97:17 ($<10^{-15}$) & 94:15 ($<10^{-15}$) & 80:11 ($<10^{-14}$) & 48:6 ($<10^{-9}$) & 4:0 ($0.06$) \\
\bottomrule
\end{tabular}
}
\end{table}

\begin{table}[H]
\centering
\caption{Sensitivity analysis: Overall (all 4 families) -- Base vs.\ Aligned wins across mass and min-corr thresholds}
\label{tab:sensitivity_overall}
\resizebox{\columnwidth}{!}{%
\begin{tabular}{l c c c c c c}
\toprule
Mass $\downarrow$ / Min-Corr $\rightarrow$ & None & 0.1 & 0.2 & 0.3 & 0.4 & 0.5 \\
\midrule
None & 380:100 ($<10^{-40}$) & 350:67 ($<10^{-47}$) & 296:57 ($<10^{-40}$) & 241:27 ($<10^{-44}$) & 89:10 ($<10^{-17}$) & 4:0 ($0.06$) \\
0.5 & 336:83 ($<10^{-37}$) & 315:64 ($<10^{-41}$) & 273:56 ($<10^{-35}$) & 224:26 ($<10^{-41}$) & 89:10 ($<10^{-17}$) & 4:0 ($0.06$) \\
0.6 & 329:80 ($<10^{-37}$) & 311:63 ($<10^{-41}$) & 270:55 ($<10^{-35}$) & 222:25 ($<10^{-41}$) & 89:9 ($<10^{-18}$) & 4:0 ($0.06$) \\
0.7 & 311:65 ($<10^{-40}$) & 297:51 ($<10^{-43}$) & 268:46 ($<10^{-39}$) & 220:22 ($<10^{-42}$) & 89:9 ($<10^{-18}$) & 4:0 ($0.06$) \\
0.8 & 284:50 ($<10^{-41}$) & 275:43 ($<10^{-43}$) & 257:40 ($<10^{-40}$) & \textbf{213:22 ($<10^{-41}$)} & 88:9 ($<10^{-17}$) & 4:0 ($0.06$) \\
0.9 & 249:25 ($<10^{-48}$) & 245:24 ($<10^{-47}$) & 231:22 ($<10^{-45}$) & 197:17 ($<10^{-40}$) & 85:7 ($<10^{-18}$) & 4:0 ($0.06$) \\
\bottomrule
\end{tabular}
}
\end{table}



\begin{footnotesize}
\begin{longtable}{r | r r r r | r r r r | r r r r}
\caption{Per-pair prediction results: Bargaining, Persuasion, and Negotiation. Pair numbers correspond to Appendix~\ref{app:models}.}\label{tab:perpair_bpn} \\
\toprule
\multicolumn{1}{c}{}  & \multicolumn{4}{c|}{Bargaining} & \multicolumn{4}{c|}{Persuasion} & \multicolumn{4}{c}{Negotiation} \\
\# & Mass$_B$ & Mass$_A$ & Corr$_B$ & Corr$_A$ & Mass$_B$ & Mass$_A$ & Corr$_B$ & Corr$_A$ & Mass$_B$ & Mass$_A$ & Corr$_B$ & Corr$_A$ \\
\midrule
\endfirsthead
\multicolumn{13}{c}{\textit{(continued)}} \\
\toprule
\multicolumn{1}{c}{}  & \multicolumn{4}{c|}{Bargaining} & \multicolumn{4}{c|}{Persuasion} & \multicolumn{4}{c}{Negotiation} \\
\# & Mass$_B$ & Mass$_A$ & Corr$_B$ & Corr$_A$ & Mass$_B$ & Mass$_A$ & Corr$_B$ & Corr$_A$ & Mass$_B$ & Mass$_A$ & Corr$_B$ & Corr$_A$ \\
\midrule
\endhead
\midrule
\multicolumn{13}{r}{\textit{continued on next page}} \\
\endfoot
\bottomrule
\endlastfoot
1 & 0.99 & 1.00 & 0.26 & -0.04 & 0.74 & 0.01 & 0.11 & -0.05 & 0.90 & 0.95 & 0.16 & 0.04 \\
2 & 0.81 & 0.79 & -0.02 & 0.01 & 0.91 & 0.02 & 0.36 & -0.07 & 0.71 & 0.42 & -0.09 & -0.01 \\
3 & 0.79 & 0.79 & -0.29 & 0.01 & 0.91 & 0.02 & 0.37 & -0.07 & 0.65 & 0.42 & -0.06 & -0.01 \\
4 & 0.75 & 0.82 & -0.20 & -0.06 & 0.91 & 0.02 & 0.31 & -0.04 & 0.76 & 0.32 & 0.05 & 0.02 \\
5 & 0.84 & 0.82 & -0.27 & -0.06 & 0.94 & 0.02 & 0.36 & -0.04 & 0.80 & 0.32 & 0.00 & 0.02 \\
6 & 0.61 & 0.28 & -0.12 & 0.03 & 0.94 & 0.05 & 0.37 & 0.09 & 0.58 & 0.36 & -0.04 & 0.05 \\
7 & 0.68 & 0.28 & 0.27 & 0.03 & 0.94 & 0.05 & 0.39 & 0.09 & 0.65 & 0.36 & 0.27 & 0.05 \\
8 & 0.90 & 0.90 & 0.11 & 0.21 & 0.67 & 0.94 & 0.07 & 0.21 & 0.73 & 0.70 & 0.32 & 0.07 \\
9 & 0.85 & 0.60 & 0.10 & 0.01 & 0.68 & 0.94 & 0.14 & 0.34 & 0.65 & 0.43 & 0.30 & 0.22 \\
10 & 0.96 & 1.00 & 0.37 & -0.06 & 0.70 & 0.88 & 0.08 & 0.25 & 0.78 & 0.80 & -0.31 & -0.13 \\
11 & 0.97 & 0.99 & 0.19 & -0.06 & 0.69 & 0.94 & 0.11 & 0.26 & 0.83 & 0.92 & -0.37 & -0.28 \\
12 & 0.76 & 0.41 & -0.29 & -0.08 & 0.93 & 0.94 & 0.35 & 0.29 & 0.58 & 0.52 & 0.31 & 0.30 \\
13 & 0.99 & 0.99 & 0.43 & -0.03 & 0.69 & 0.64 & 0.12 & 0.11 & 0.77 & 0.90 & -0.21 & -0.26 \\
14 & 0.68 & 0.44 & 0.03 & 0.10 & 0.71 & 0.59 & 0.12 & 0.10 & 0.58 & 0.43 & 0.26 & 0.03 \\
15 & 0.97 & 1.00 & 0.45 & 0.42 & 0.95 & 0.94 & 0.32 & 0.27 & 0.82 & 0.95 & 0.27 & 0.14 \\
16 & 0.83 & 0.65 & 0.08 & 0.11 & 0.72 & 0.64 & 0.10 & 0.05 & 0.41 & 0.25 & -0.06 & -0.01 \\
17 & 0.99 & 1.00 & 0.37 & 0.15 & 0.94 & 0.95 & 0.36 & 0.27 & 0.95 & 1.00 & 0.34 & 0.24 \\
18 & 0.70 & 0.21 & 0.10 & 0.10 & 0.69 & 0.62 & 0.15 & 0.06 & 0.36 & 0.51 & -0.07 & -0.06 \\
19 & 0.65 & 0.52 & -0.15 & -0.01 & 0.93 & 0.92 & 0.35 & 0.35 & 0.43 & 0.20 & -0.14 & -0.05 \\
20 & 0.99 & 1.00 & 0.44 & 0.08 & 0.66 & 0.95 & 0.15 & 0.23 & 0.89 & 0.98 & 0.17 & 0.06 \\
21 & 0.67 & 0.36 & -0.01 & -0.06 & 0.93 & 0.93 & 0.33 & 0.31 & 0.50 & 0.33 & 0.01 & 0.01 \\
22 & 0.98 & 0.99 & 0.33 & -0.22 & 0.72 & 0.62 & 0.18 & 0.11 & 0.85 & 0.94 & -0.22 & -0.14 \\
23 & 0.99 & 1.00 & 0.39 & -0.05 & 0.93 & 0.95 & 0.36 & 0.25 & 0.80 & 1.00 & 0.06 & -0.07 \\
24 & 0.98 & 0.99 & 0.32 & -0.32 & 0.73 & 0.63 & 0.19 & 0.11 & 0.88 & 1.00 & 0.26 & -0.05 \\
25 & 0.96 & 1.00 & 0.45 & -0.01 & 0.75 & 0.25 & 0.10 & -0.04 & 0.96 & 1.00 & 0.27 & 0.06 \\
26 & 0.91 & 1.00 & 0.43 & 0.11 & 0.76 & 0.07 & 0.12 & 0.01 & 0.91 & 0.97 & 0.36 & -0.01 \\
27 & 0.89 & 1.00 & 0.44 & 0.02 & 0.76 & 0.00 & 0.12 & -0.04 & 0.92 & 1.00 & 0.19 & -0.05 \\
28 & 0.95 & 1.00 & 0.30 & 0.05 & 0.72 & 0.00 & 0.12 & -0.04 & 0.89 & 1.00 & -0.14 & -0.03 \\
29 & 0.99 & 1.00 & 0.41 & 0.07 & 0.73 & 0.01 & 0.12 & -0.07 & 0.91 & 1.00 & 0.32 & -0.01 \\
30 & 0.99 & 1.00 & 0.35 & -0.00 & 0.74 & 0.02 & 0.14 & -0.04 & 0.93 & 0.88 & 0.22 & 0.03 \\
31 & 0.99 & 1.00 & 0.38 & 0.11 & 0.70 & 0.01 & 0.15 & 0.13 & 0.90 & 0.99 & 0.05 & 0.03 \\
32 & 0.97 & 1.00 & 0.30 & -0.01 & 0.94 & 0.00 & 0.32 & -0.04 & 0.80 & 1.00 & 0.35 & 0.01 \\
33 & 0.98 & 1.00 & 0.35 & 0.00 & 0.93 & 0.01 & 0.35 & 0.04 & 0.91 & 1.00 & 0.26 & -0.02 \\
34 & 0.99 & 1.00 & 0.44 & 0.13 & 0.94 & 0.00 & 0.32 & 0.08 & 0.96 & 1.00 & 0.07 & -0.04 \\
35 & 0.99 & 1.00 & 0.42 & 0.08 & 0.92 & 0.00 & 0.36 & 0.02 & 0.96 & 1.00 & 0.38 & 0.04 \\
36 & 0.97 & 1.00 & 0.41 & 0.27 & 0.95 & 0.95 & 0.35 & 0.25 & 0.95 & 0.99 & 0.38 & -0.10 \\
37 & 0.96 & 1.00 & 0.42 & 0.16 & 0.95 & 0.95 & 0.34 & 0.30 & 0.95 & 1.00 & 0.37 & -0.12 \\
38 & 0.99 & 1.00 & 0.11 & 0.01 & 0.95 & 0.95 & 0.36 & 0.23 & 0.82 & 0.77 & 0.32 & -0.30 \\
39 & 0.99 & 1.00 & 0.16 & -0.10 & 0.95 & 0.95 & 0.36 & 0.31 & 0.78 & 0.86 & 0.09 & -0.32 \\
40 & 0.99 & 1.00 & 0.44 & 0.21 & 0.94 & 0.95 & 0.32 & 0.26 & 0.94 & 0.99 & 0.41 & -0.15 \\
41 & 0.74 & 0.82 & 0.42 & 0.05 & 0.97 & 0.37 & 0.11 & -0.03 & 0.76 & 0.90 & -0.24 & -0.00 \\
42 & 0.58 & 0.99 & 0.34 & 0.44 & 0.64 & 0.95 & 0.08 & 0.17 & 0.05 & 0.91 & -0.19 & -0.16 \\
43 & 0.97 & 0.99 & 0.38 & 0.03 & 0.68 & 0.02 & 0.11 & 0.07 & 0.76 & 0.89 & -0.33 & 0.04 \\
44 & 0.97 & 0.99 & 0.42 & -0.21 & 0.86 & 0.90 & 0.15 & 0.22 & 0.87 & 0.96 & 0.28 & -0.18 \\
45 & 0.98 & 1.00 & 0.43 & -0.13 & 0.84 & 0.92 & 0.20 & 0.26 & 0.90 & 0.99 & 0.26 & -0.04 \\
46 & 0.58 & 0.41 & 0.11 & -0.00 & 0.65 & 0.00 & 0.11 & -0.04 & 0.54 & 0.45 & -0.19 & 0.04 \\
47 & 0.99 & 1.00 & 0.45 & -0.03 & 0.85 & 0.94 & 0.21 & 0.22 & 0.85 & 1.00 & 0.35 & -0.02 \\
48 & 0.99 & 1.00 & 0.45 & -0.22 & 0.84 & 0.95 & 0.20 & 0.23 & 0.82 & 1.00 & 0.34 & -0.04 \\
49 & 0.72 & 0.62 & 0.10 & -0.02 & 0.91 & 0.00 & 0.37 & -0.04 & 0.51 & 0.03 & 0.07 & 0.05 \\
50 & 0.99 & 1.00 & 0.35 & -0.36 & 0.84 & 0.71 & 0.20 & 0.07 & 0.86 & 1.00 & 0.10 & -0.06 \\
51 & 1.00 & 1.00 & 0.39 & -0.27 & 0.84 & 0.72 & 0.20 & 0.10 & 0.90 & 1.00 & 0.11 & -0.09 \\
52 & 0.97 & 0.75 & 0.35 & 0.20 & 0.94 & 0.96 & 0.30 & 0.20 & 0.92 & 0.14 & 0.34 & -0.12 \\
53 & 0.98 & 1.00 & 0.43 & 0.18 & 0.95 & 0.94 & 0.36 & 0.31 & 0.90 & 0.99 & 0.35 & 0.22 \\
54 & 0.98 & 1.00 & 0.43 & 0.09 & 0.67 & 0.00 & 0.10 & -0.07 & 0.76 & 0.79 & 0.11 & 0.03 \\
55 & 0.98 & 1.00 & 0.43 & 0.06 & 0.67 & 0.00 & 0.09 & 0.00 & 0.73 & 0.68 & 0.12 & 0.00 \\
56 & 0.98 & 0.99 & 0.43 & 0.03 & 0.67 & 0.22 & 0.09 & -0.08 & 0.73 & 0.76 & 0.12 & 0.08 \\
57 & 0.99 & 1.00 & 0.40 & 0.06 & 0.95 & 0.01 & 0.36 & -0.03 & 0.80 & 0.93 & 0.13 & 0.00 \\
58 & 0.99 & 1.00 & 0.32 & -0.31 & 0.78 & 0.71 & 0.15 & 0.13 & 0.88 & 0.92 & -0.10 & -0.21 \\
59 & 0.98 & 1.00 & 0.38 & -0.23 & 0.94 & 0.91 & 0.36 & 0.25 & 0.94 & 0.99 & 0.04 & -0.39 \\
60 & 0.96 & 1.00 & 0.31 & 0.35 & 0.94 & 0.93 & 0.33 & 0.16 & 0.95 & 0.99 & 0.23 & 0.14 \\
61 & 0.00 & 0.96 & 0.39 & 0.08 & 0.00 & 0.98 & -0.05 & 0.00 & 0.00 & 0.45 & -0.02 & -0.11 \\
62 & 0.97 & 1.00 & 0.29 & -0.12 & 0.90 & 0.84 & 0.37 & 0.28 & 0.92 & 0.71 & 0.30 & 0.08 \\
63 & 0.97 & 1.00 & 0.34 & -0.10 & 0.91 & 0.50 & 0.35 & 0.15 & 0.84 & 0.92 & 0.17 & -0.19 \\
64 & 0.95 & 1.00 & 0.43 & 0.43 & 0.95 & 0.91 & 0.31 & 0.20 & 0.83 & 0.84 & -0.19 & -0.15 \\
65 & 0.99 & 1.00 & 0.22 & -0.25 & 0.80 & 0.74 & 0.24 & 0.20 & 0.94 & 1.00 & 0.30 & -0.12 \\
66 & 0.99 & 1.00 & 0.19 & -0.23 & 0.83 & 0.64 & 0.19 & 0.16 & 0.94 & 1.00 & 0.17 & -0.23 \\
67 & 0.99 & 1.00 & 0.44 & -0.14 & 0.77 & 0.72 & 0.18 & 0.05 & 0.94 & 0.99 & 0.30 & 0.04 \\
68 & 0.95 & 0.99 & 0.34 & 0.14 & 0.94 & 0.92 & 0.35 & 0.31 & 0.91 & 0.99 & 0.30 & -0.09 \\
69 & 0.80 & 0.90 & 0.42 & -0.15 & 0.95 & 0.86 & 0.31 & 0.15 & 0.88 & 0.96 & 0.27 & -0.13 \\
70 & 0.93 & 0.99 & 0.38 & 0.27 & 0.96 & 0.95 & 0.23 & 0.19 & 0.94 & 0.98 & 0.25 & -0.12 \\
71 & 0.98 & 0.98 & 0.43 & 0.18 & 0.82 & 0.69 & 0.17 & 0.09 & 0.93 & 0.97 & 0.42 & -0.17 \\
72 & 0.99 & 1.00 & 0.40 & 0.11 & 0.95 & 0.92 & 0.34 & 0.31 & 0.94 & 0.99 & 0.32 & 0.40 \\
73 & 0.99 & 1.00 & 0.38 & 0.23 & 0.81 & 0.94 & 0.17 & 0.32 & 0.94 & 0.99 & 0.39 & -0.01 \\
74 & 0.97 & 1.00 & 0.35 & 0.28 & 0.81 & 0.91 & 0.15 & 0.29 & 0.92 & 0.98 & 0.30 & 0.03 \\
75 & 0.94 & 0.33 & 0.38 & 0.24 & 0.93 & 0.94 & 0.20 & 0.23 & 0.84 & 0.37 & 0.24 & 0.11 \\
76 & 0.98 & 0.99 & 0.44 & 0.40 & 0.82 & 0.70 & 0.13 & 0.06 & 0.93 & 0.98 & 0.37 & -0.21 \\
77 & 0.97 & 1.00 & 0.42 & 0.05 & 0.79 & 0.87 & 0.17 & 0.20 & 0.92 & 1.00 & 0.41 & 0.26 \\
78 & 0.96 & 1.00 & 0.41 & 0.36 & 0.82 & 0.90 & 0.15 & 0.26 & 0.97 & 1.00 & 0.31 & 0.12 \\
79 & 0.94 & 1.00 & 0.30 & -0.08 & 0.81 & 0.94 & 0.14 & 0.31 & 0.86 & 1.00 & 0.32 & 0.15 \\
80 & 0.99 & 0.99 & 0.36 & 0.30 & 0.80 & 0.69 & 0.17 & 0.10 & 0.88 & 0.99 & 0.16 & -0.03 \\
81 & 0.98 & 1.00 & 0.37 & 0.08 & 0.82 & 0.95 & 0.12 & 0.31 & 0.88 & 0.97 & 0.12 & 0.02 \\
82 & 0.99 & 1.00 & 0.42 & 0.11 & 0.81 & 0.95 & 0.20 & 0.28 & 0.82 & 1.00 & 0.28 & 0.06 \\
83 & 0.99 & 1.00 & 0.42 & -0.07 & 0.81 & 0.68 & 0.20 & 0.13 & 0.82 & 1.00 & 0.28 & -0.03 \\
84 & 0.99 & 1.00 & 0.41 & -0.10 & 0.79 & 0.94 & 0.20 & 0.22 & 0.78 & 1.00 & 0.26 & -0.03 \\
85 & 0.99 & 1.00 & 0.36 & 0.05 & 0.81 & 0.94 & 0.16 & 0.26 & 0.82 & 0.98 & 0.34 & 0.03 \\
86 & 0.98 & 1.00 & 0.34 & -0.16 & 0.82 & 0.89 & 0.21 & 0.27 & 0.91 & 1.00 & 0.30 & -0.19 \\
87 & 0.99 & 0.09 & 0.29 & 0.01 & 0.90 & 0.45 & 0.31 & 0.35 & 0.88 & 0.12 & 0.32 & -0.02 \\
88 & 0.99 & 0.99 & 0.42 & 0.27 & 0.81 & 0.70 & 0.14 & 0.09 & 0.81 & 0.98 & 0.36 & 0.15 \\
89 & 0.99 & 1.00 & 0.34 & 0.06 & 0.80 & 0.94 & 0.20 & 0.28 & 0.84 & 1.00 & 0.30 & 0.08 \\
90 & 0.99 & 1.00 & 0.13 & -0.11 & 0.94 & 0.90 & 0.35 & 0.18 & 0.94 & 1.00 & 0.28 & -0.10 \\
91 & 0.99 & 0.99 & 0.36 & -0.22 & 0.94 & 0.67 & 0.32 & 0.11 & 0.84 & 0.97 & 0.35 & -0.03 \\
92 & 0.98 & 1.00 & 0.35 & 0.08 & 0.93 & 0.92 & 0.37 & 0.30 & 0.86 & 0.80 & 0.32 & 0.16 \\
93 & 0.99 & 1.00 & 0.39 & 0.04 & 0.94 & 0.94 & 0.34 & 0.31 & 0.81 & 1.00 & 0.23 & -0.01 \\
94 & 0.99 & 0.99 & 0.25 & -0.11 & 0.80 & 0.70 & 0.20 & 0.10 & 0.95 & 0.99 & 0.37 & -0.17 \\
95 & 0.99 & 1.00 & 0.33 & -0.04 & 0.93 & 0.94 & 0.33 & 0.24 & 0.94 & 1.00 & 0.19 & -0.11 \\
96 & 0.99 & 1.00 & 0.40 & 0.06 & 0.91 & 0.92 & 0.34 & 0.23 & 0.80 & 1.00 & 0.35 & -0.11 \\
97 & 1.00 & 1.00 & 0.36 & -0.02 & 0.94 & 0.93 & 0.34 & 0.27 & 0.92 & 1.00 & 0.39 & -0.12 \\
98 & 0.99 & 1.00 & 0.31 & -0.05 & 0.79 & 0.69 & 0.18 & 0.13 & 0.93 & 1.00 & 0.30 & -0.08 \\
99 & 0.99 & 1.00 & 0.23 & -0.07 & 0.81 & 0.70 & 0.18 & 0.13 & 0.96 & 1.00 & 0.24 & -0.13 \\
100 & 0.93 & 0.99 & 0.39 & 0.36 & 0.93 & 0.83 & 0.33 & 0.25 & 0.81 & 0.98 & 0.10 & 0.11 \\
101 & 0.93 & 0.99 & 0.30 & 0.08 & 0.93 & 0.89 & 0.35 & 0.28 & 0.85 & 0.97 & 0.33 & -0.16 \\
102 & 0.99 & 1.00 & 0.40 & 0.10 & 0.95 & 0.94 & 0.31 & 0.29 & 0.72 & 0.98 & -0.13 & -0.05 \\
103 & 0.72 & 0.83 & 0.08 & 0.04 & 0.76 & 0.70 & 0.08 & 0.19 & 0.77 & 0.50 & 0.28 & 0.03 \\
104 & 0.62 & 0.66 & 0.29 & 0.31 & 0.86 & 0.77 & 0.23 & 0.09 & 0.77 & 0.79 & 0.14 & 0.18 \\
105 & 0.72 & 0.64 & 0.21 & 0.23 & 0.97 & 0.95 & 0.31 & 0.23 & 0.69 & 0.70 & -0.11 & -0.09 \\
106 & 0.83 & 0.79 & 0.10 & 0.07 & 0.88 & 0.75 & 0.29 & 0.11 & 0.83 & 0.82 & 0.33 & -0.01 \\
107 & 0.66 & 0.38 & 0.26 & 0.23 & 0.94 & 0.95 & 0.26 & 0.22 & 0.72 & 0.46 & 0.06 & -0.02 \\
108 & 0.59 & 0.77 & 0.04 & 0.07 & 0.76 & 0.74 & 0.10 & 0.13 & 0.80 & 0.92 & -0.06 & 0.29 \\
109 & 0.72 & 0.65 & 0.04 & 0.07 & 0.76 & 0.95 & 0.11 & 0.29 & 0.64 & 0.38 & 0.16 & 0.06 \\
110 & 0.99 & 0.94 & 0.32 & -0.05 & 0.93 & 0.95 & 0.34 & 0.23 & 0.78 & 0.72 & -0.24 & -0.24 \\
111 & 0.98 & 1.00 & 0.41 & 0.19 & 0.83 & 0.94 & 0.10 & 0.28 & 0.95 & 1.00 & 0.15 & 0.02 \\
112 & 0.97 & 1.00 & 0.32 & 0.34 & 0.80 & 0.92 & 0.17 & 0.20 & 0.83 & 0.88 & -0.09 & -0.03 \\
113 & 0.99 & 1.00 & 0.32 & 0.08 & 0.93 & 0.95 & 0.37 & 0.23 & 0.91 & 0.83 & 0.20 & -0.07 \\
114 & 0.77 & 0.85 & 0.11 & 0.07 & 0.66 & 0.62 & 0.05 & 0.04 & 0.62 & 0.65 & 0.06 & -0.32 \\
115 & 0.96 & 1.00 & 0.42 & 0.00 & 0.66 & 0.95 & 0.15 & 0.23 & 0.59 & 0.90 & 0.08 & -0.33 \\
116 & 0.95 & 1.00 & 0.31 & 0.37 & 0.68 & 0.95 & 0.10 & 0.31 & 0.70 & 0.82 & -0.32 & -0.00 \\
117 & 0.97 & 1.00 & 0.45 & 0.07 & 0.68 & 0.95 & 0.14 & 0.28 & 0.80 & 0.89 & -0.05 & -0.33 \\
118 & 0.97 & 1.00 & 0.42 & 0.22 & 0.94 & 0.94 & 0.37 & 0.30 & 0.85 & 0.87 & -0.18 & -0.19 \\
119 & 0.97 & 1.00 & 0.42 & 0.02 & 0.94 & 0.95 & 0.38 & 0.31 & 0.87 & 0.92 & -0.29 & -0.32 \\
120 & 0.93 & 0.96 & 0.40 & 0.35 & 0.68 & 0.58 & 0.05 & 0.05 & 0.76 & 0.70 & -0.01 & 0.23 \\
\end{longtable}
\end{footnotesize}

\begin{footnotesize}
\begin{longtable}{r | r r r r | r r r r | r r r r}
\caption{Per-pair prediction results: Repeated Matrix Games, One-Shot $2\times2$ Games, and Binary Lotteries. Pair numbers correspond to Appendix~\ref{app:models}.}\label{tab:perpair_matrix_boundary} \\
\toprule
\multicolumn{1}{c}{}  & \multicolumn{4}{c|}{Matrix} & \multicolumn{4}{c|}{One-Shot $2{\times}2$} & \multicolumn{4}{c}{Lotteries} \\
\# & Mass$_B$ & Mass$_A$ & Corr$_B$ & Corr$_A$ & Mass$_B$ & Mass$_A$ & Corr$_B$ & Corr$_A$ & Mass$_B$ & Mass$_A$ & Corr$_B$ & Corr$_A$ \\
\midrule
\endfirsthead
\multicolumn{13}{c}{\textit{(continued)}} \\
\toprule
\multicolumn{1}{c}{}  & \multicolumn{4}{c|}{Matrix} & \multicolumn{4}{c|}{One-Shot $2{\times}2$} & \multicolumn{4}{c}{Lotteries} \\
\# & Mass$_B$ & Mass$_A$ & Corr$_B$ & Corr$_A$ & Mass$_B$ & Mass$_A$ & Corr$_B$ & Corr$_A$ & Mass$_B$ & Mass$_A$ & Corr$_B$ & Corr$_A$ \\
\midrule
\endhead
\midrule
\multicolumn{13}{r}{\textit{continued on next page}} \\
\endfoot
\bottomrule
\endlastfoot
1 & 0.99 & 1.00 & 0.41 & 0.32 & 0.81 & 1.00 & -0.02 & -0.05 & 0.93 & 1.00 & 0.10 & 0.45 \\
2 & 0.99 & 1.00 & 0.35 & -0.05 & 0.88 & 0.99 & -0.00 & 0.05 & 0.98 & 1.00 & 0.29 & 0.55 \\
3 & 0.99 & 1.00 & 0.39 & -0.05 & 0.88 & 0.99 & -0.01 & 0.05 & 0.97 & 1.00 & 0.22 & 0.55 \\
4 & 1.00 & 1.00 & 0.43 & 0.18 & 0.82 & 0.98 & -0.02 & -0.00 & 0.96 & 1.00 & 0.29 & 0.67 \\
5 & 0.99 & 1.00 & 0.36 & 0.18 & 0.82 & 0.98 & 0.06 & -0.00 & 0.96 & 1.00 & 0.31 & 0.67 \\
6 & 1.00 & 1.00 & 0.41 & -0.06 & 0.93 & 0.99 & 0.03 & 0.12 & 0.97 & 1.00 & 0.42 & 0.75 \\
7 & 1.00 & 1.00 & 0.41 & -0.06 & 0.95 & 0.99 & 0.05 & 0.12 & 0.95 & 1.00 & 0.25 & 0.75 \\
8 & 0.99 & 1.00 & 0.35 & 0.43 & 0.96 & 0.99 & -0.01 & 0.00 & 0.98 & 1.00 & 0.01 & 0.30 \\
9 & 0.99 & 1.00 & 0.47 & 0.34 & 0.74 & 0.98 & -0.00 & -0.02 & 0.95 & 1.00 & 0.28 & 0.66 \\
10 & 0.99 & 1.00 & 0.28 & 0.05 & 0.92 & 0.90 & 0.04 & 0.08 & 0.98 & 1.00 & 0.34 & 0.67 \\
11 & 1.00 & 1.00 & 0.39 & 0.35 & 0.95 & 0.99 & 0.01 & -0.02 & 0.99 & 0.99 & 0.49 & 0.72 \\
12 & 0.99 & 1.00 & 0.39 & 0.37 & 0.73 & 0.79 & 0.03 & -0.05 & 0.93 & 0.99 & 0.37 & 0.59 \\
13 & 1.00 & 1.00 & 0.44 & 0.44 & 0.92 & 0.99 & 0.03 & -0.05 & 0.98 & 1.00 & 0.65 & 0.77 \\
14 & 0.99 & 1.00 & 0.19 & 0.13 & 0.92 & 1.00 & 0.01 & 0.02 & 0.99 & 1.00 & 0.39 & 0.62 \\
15 & 0.98 & 1.00 & 0.24 & 0.43 & 0.96 & 0.99 & -0.04 & -0.01 & 0.98 & 0.97 & 0.40 & 0.35 \\
16 & 0.99 & 1.00 & 0.30 & -0.01 & 0.97 & 1.00 & -0.03 & 0.06 & 0.99 & 1.00 & 0.68 & 0.59 \\
17 & 0.98 & 1.00 & 0.37 & 0.14 & 0.96 & 1.00 & 0.04 & -0.06 & 0.97 & 1.00 & 0.68 & 0.56 \\
18 & 0.99 & 1.00 & 0.40 & 0.16 & 0.94 & 1.00 & 0.00 & 0.02 & 0.94 & 1.00 & 0.68 & 0.69 \\
19 & 0.99 & 1.00 & 0.20 & 0.27 & 0.91 & 0.84 & -0.00 & 0.02 & 0.98 & 0.99 & 0.36 & 0.71 \\
20 & 0.99 & 1.00 & 0.33 & 0.25 & 0.96 & 1.00 & -0.02 & -0.13 & 0.98 & 0.99 & 0.69 & 0.75 \\
21 & 0.99 & 1.00 & 0.40 & 0.41 & 0.85 & 0.88 & 0.04 & 0.01 & 0.99 & 1.00 & 0.58 & 0.68 \\
22 & 0.99 & 1.00 & 0.41 & 0.28 & 0.96 & 1.00 & -0.00 & 0.04 & 0.96 & 1.00 & 0.56 & 0.73 \\
23 & 0.99 & 1.00 & 0.41 & 0.33 & 0.98 & 1.00 & -0.02 & 0.08 & 0.99 & 1.00 & 0.74 & 0.75 \\
24 & 1.00 & 1.00 & 0.48 & 0.21 & 0.97 & 1.00 & -0.02 & -0.02 & 0.99 & 1.00 & 0.79 & 0.76 \\
25 & 0.99 & 1.00 & 0.28 & 0.17 & 0.93 & 1.00 & -0.06 & -0.04 & 0.96 & 1.00 & 0.03 & 0.29 \\
26 & 0.98 & 1.00 & 0.30 & 0.08 & 0.90 & 0.98 & 0.04 & 0.04 & 0.98 & 0.99 & 0.20 & 0.48 \\
27 & 0.99 & 1.00 & 0.39 & -0.34 & 0.98 & 1.00 & 0.05 & 0.05 & 0.97 & 1.00 & 0.22 & 0.43 \\
28 & 0.99 & 1.00 & 0.18 & -0.09 & 0.94 & 1.00 & 0.00 & 0.03 & 0.98 & 1.00 & 0.04 & 0.02 \\
29 & 0.99 & 1.00 & 0.37 & 0.12 & 0.95 & 1.00 & -0.00 & 0.04 & 0.98 & 1.00 & 0.24 & 0.59 \\
30 & 0.99 & 1.00 & 0.44 & -0.08 & 0.94 & 1.00 & 0.01 & 0.02 & 0.97 & 1.00 & 0.12 & 0.61 \\
31 & 1.00 & 1.00 & 0.48 & 0.24 & 0.93 & 1.00 & 0.01 & -0.13 & 0.97 & 0.97 & 0.25 & 0.67 \\
32 & 0.99 & 1.00 & -0.09 & 0.23 & 0.86 & 1.00 & 0.04 & -0.05 & 0.93 & 1.00 & 0.09 & 0.52 \\
33 & 1.00 & 1.00 & 0.48 & 0.24 & 0.95 & 1.00 & -0.01 & 0.01 & 0.95 & 1.00 & 0.44 & 0.67 \\
34 & 1.00 & 1.00 & 0.47 & 0.36 & 0.97 & 1.00 & -0.03 & -0.01 & 0.97 & 1.00 & 0.56 & 0.70 \\
35 & 1.00 & 1.00 & 0.47 & 0.40 & 0.90 & 1.00 & -0.02 & 0.01 & 0.96 & 1.00 & 0.46 & 0.69 \\
36 & 1.00 & 1.00 & 0.37 & 0.26 & 0.98 & 1.00 & -0.05 & -0.04 & 0.99 & 1.00 & 0.52 & 0.67 \\
37 & 1.00 & 1.00 & 0.36 & 0.29 & 0.98 & 0.99 & -0.07 & -0.03 & 1.00 & 1.00 & 0.56 & 0.60 \\
38 & 1.00 & 1.00 & 0.42 & 0.34 & 1.00 & 0.99 & 0.04 & -0.07 & 1.00 & 1.00 & 0.75 & 0.72 \\
39 & 1.00 & 1.00 & 0.44 & 0.32 & 1.00 & 1.00 & 0.01 & 0.03 & 1.00 & 1.00 & 0.74 & 0.67 \\
40 & 1.00 & 1.00 & 0.49 & 0.42 & 0.94 & 1.00 & 0.05 & 0.14 & 0.94 & 1.00 & 0.57 & 0.62 \\
41 & 0.98 & 0.94 & 0.25 & 0.31 & 0.93 & 0.95 & -0.01 & -0.02 & 0.93 & 0.94 & 0.09 & 0.18 \\
42 & 0.86 & 1.00 & 0.25 & 0.48 & 0.66 & 1.00 & -0.02 & -0.02 & 0.89 & 1.00 & 0.05 & 0.57 \\
43 & 0.99 & 1.00 & 0.05 & 0.34 & 0.97 & 0.92 & 0.03 & -0.03 & 0.99 & 1.00 & 0.46 & 0.53 \\
44 & 0.99 & 1.00 & 0.15 & 0.18 & 0.93 & 1.00 & -0.04 & -0.02 & 0.98 & 1.00 & 0.33 & 0.50 \\
45 & 1.00 & 1.00 & 0.28 & -0.15 & 0.96 & 0.94 & 0.04 & 0.03 & 0.98 & 1.00 & 0.35 & 0.55 \\
46 & 1.00 & 1.00 & 0.23 & -0.29 & 0.95 & 1.00 & 0.04 & 0.06 & 0.98 & 1.00 & 0.32 & 0.63 \\
47 & 1.00 & 1.00 & 0.41 & -0.15 & 0.94 & 1.00 & 0.01 & 0.14 & 0.96 & 1.00 & 0.19 & 0.67 \\
48 & 1.00 & 1.00 & 0.42 & -0.21 & 0.95 & 1.00 & -0.01 & 0.06 & 0.97 & 1.00 & 0.24 & 0.73 \\
49 & 0.99 & 1.00 & 0.30 & -0.30 & 0.97 & 1.00 & -0.01 & 0.08 & 0.98 & 1.00 & 0.26 & 0.58 \\
50 & 1.00 & 1.00 & 0.51 & 0.02 & 0.95 & 1.00 & -0.03 & 0.01 & 0.97 & 1.00 & 0.35 & 0.75 \\
51 & 1.00 & 1.00 & 0.51 & 0.06 & 0.96 & 1.00 & -0.00 & -0.03 & 0.98 & 1.00 & 0.57 & 0.77 \\
52 & 0.99 & 0.84 & 0.42 & 0.23 & 0.92 & 0.00 & 0.05 & 0.01 & 0.91 & 0.01 & 0.49 & 0.34 \\
53 & 1.00 & 1.00 & 0.45 & 0.42 & 0.95 & 1.00 & 0.03 & -0.04 & 0.96 & 1.00 & 0.70 & 0.57 \\
54 & 0.99 & 1.00 & 0.38 & 0.10 & 0.95 & 0.97 & -0.06 & 0.15 & 0.95 & 0.76 & 0.30 & 0.57 \\
55 & 0.99 & 1.00 & 0.41 & 0.11 & 0.88 & 0.77 & -0.03 & 0.12 & 0.96 & 0.82 & 0.39 & 0.56 \\
56 & 0.99 & 1.00 & 0.41 & -0.17 & 0.88 & 0.98 & -0.03 & -0.05 & 0.96 & 0.98 & 0.39 & 0.71 \\
57 & 1.00 & 1.00 & 0.45 & 0.09 & 0.98 & 1.00 & -0.03 & -0.12 & 0.98 & 1.00 & 0.05 & 0.67 \\
58 & 0.99 & 1.00 & 0.48 & 0.27 & 0.93 & 1.00 & 0.01 & 0.02 & 0.95 & 1.00 & 0.64 & 0.78 \\
59 & 0.99 & 1.00 & 0.12 & -0.26 & 0.86 & 0.91 & 0.03 & -0.01 & 0.98 & 1.00 & 0.46 & 0.53 \\
60 & 0.99 & 1.00 & 0.15 & 0.38 & 0.97 & 1.00 & 0.01 & 0.01 & 0.99 & 0.98 & 0.32 & 0.59 \\
61 & 0.00 & 0.68 & 0.02 & 0.10 & 0.04 & 1.00 & -0.07 & 0.00 & 0.03 & 0.00 & -0.03 & 0.13 \\
62 & 0.99 & 1.00 & 0.36 & 0.26 & 0.94 & 1.00 & 0.02 & 0.13 & 0.99 & 1.00 & 0.26 & 0.58 \\
63 & 0.99 & 1.00 & 0.39 & -0.08 & 0.93 & 1.00 & -0.01 & -0.02 & 0.98 & 1.00 & 0.46 & 0.50 \\
64 & 1.00 & 1.00 & 0.32 & 0.21 & 0.98 & 1.00 & -0.01 & 0.03 & 1.00 & 1.00 & 0.49 & 0.60 \\
65 & 1.00 & 1.00 & 0.47 & 0.21 & 0.94 & 1.00 & -0.04 & 0.04 & 0.98 & 1.00 & 0.67 & 0.71 \\
66 & 0.99 & 1.00 & 0.47 & 0.47 & 0.93 & 1.00 & -0.03 & 0.08 & 0.98 & 1.00 & 0.68 & 0.80 \\
67 & 1.00 & 1.00 & 0.48 & 0.16 & 0.99 & 1.00 & -0.04 & -0.10 & 0.98 & 1.00 & 0.65 & 0.68 \\
68 & 0.99 & 0.99 & 0.37 & 0.36 & 0.94 & 1.00 & -0.01 & 0.00 & 1.00 & 1.00 & 0.30 & 0.41 \\
69 & 0.98 & 0.99 & 0.44 & 0.42 & 0.93 & 0.98 & 0.03 & -0.01 & 0.98 & 0.99 & 0.25 & 0.18 \\
70 & 0.97 & 0.94 & 0.20 & 0.22 & 0.96 & 0.99 & -0.01 & -0.01 & 0.99 & 0.99 & 0.24 & 0.23 \\
71 & 0.99 & 0.99 & 0.33 & 0.46 & 0.97 & 0.97 & 0.02 & 0.01 & 1.00 & 1.00 & 0.24 & 0.28 \\
72 & 0.99 & 1.00 & 0.31 & 0.21 & 0.96 & 1.00 & 0.03 & 0.03 & 0.99 & 1.00 & 0.47 & 0.51 \\
73 & 0.99 & 1.00 & 0.24 & 0.28 & 0.98 & 1.00 & 0.01 & -0.03 & 0.99 & 1.00 & 0.48 & 0.26 \\
74 & 0.99 & 1.00 & 0.27 & 0.06 & 0.97 & 1.00 & 0.01 & -0.06 & 0.98 & 0.99 & 0.55 & 0.40 \\
75 & 0.99 & 0.96 & 0.29 & 0.12 & 0.95 & 1.00 & -0.05 & 0.04 & 0.99 & 1.00 & 0.47 & 0.29 \\
76 & 1.00 & 1.00 & 0.44 & 0.44 & 0.96 & 0.96 & 0.05 & 0.03 & 1.00 & 1.00 & 0.22 & 0.32 \\
77 & 1.00 & 1.00 & 0.37 & 0.14 & 0.96 & 1.00 & 0.05 & -0.00 & 0.99 & 1.00 & 0.60 & 0.56 \\
78 & 0.98 & 0.99 & 0.24 & 0.14 & 0.90 & 0.99 & 0.00 & -0.01 & 0.99 & 1.00 & 0.54 & 0.63 \\
79 & 1.00 & 1.00 & 0.31 & 0.04 & 0.99 & 1.00 & -0.00 & 0.04 & 0.99 & 1.00 & 0.46 & 0.25 \\
80 & 1.00 & 1.00 & 0.40 & 0.44 & 0.91 & 0.97 & -0.00 & -0.06 & 0.99 & 1.00 & 0.67 & 0.70 \\
81 & 1.00 & 1.00 & 0.30 & 0.26 & 0.99 & 1.00 & 0.00 & 0.02 & 0.98 & 1.00 & 0.57 & 0.57 \\
82 & 1.00 & 1.00 & 0.49 & 0.40 & 0.99 & 0.98 & -0.05 & 0.05 & 1.00 & 1.00 & 0.73 & 0.65 \\
83 & 1.00 & 1.00 & 0.49 & 0.35 & 0.99 & 1.00 & -0.05 & 0.07 & 1.00 & 1.00 & 0.73 & 0.66 \\
84 & 1.00 & 1.00 & 0.40 & 0.36 & 0.99 & 1.00 & -0.01 & -0.01 & 1.00 & 1.00 & 0.75 & 0.67 \\
85 & 0.99 & 1.00 & 0.37 & 0.32 & 0.96 & 1.00 & 0.05 & 0.03 & 0.99 & 1.00 & 0.70 & 0.69 \\
86 & 0.99 & 1.00 & 0.35 & 0.21 & 0.99 & 1.00 & 0.05 & -0.03 & 1.00 & 1.00 & 0.63 & 0.59 \\
87 & 0.99 & 0.97 & 0.46 & 0.38 & 0.98 & 0.95 & -0.06 & -0.01 & 0.99 & 0.99 & 0.74 & 0.72 \\
88 & 1.00 & 1.00 & 0.38 & 0.39 & 0.91 & 0.99 & 0.05 & 0.16 & 0.99 & 1.00 & 0.69 & 0.69 \\
89 & 1.00 & 1.00 & 0.46 & 0.33 & 0.99 & 1.00 & -0.03 & -0.07 & 1.00 & 1.00 & 0.73 & 0.69 \\
90 & 1.00 & 1.00 & 0.47 & 0.39 & 0.99 & 1.00 & -0.03 & -0.11 & 0.99 & 1.00 & 0.72 & 0.72 \\
91 & 0.99 & 1.00 & 0.42 & 0.26 & 0.97 & 1.00 & -0.06 & 0.08 & 1.00 & 1.00 & 0.64 & 0.65 \\
92 & 1.00 & 1.00 & 0.34 & 0.25 & 0.96 & 1.00 & -0.01 & 0.03 & 0.99 & 1.00 & 0.53 & 0.62 \\
93 & 1.00 & 1.00 & 0.45 & 0.44 & 0.96 & 1.00 & -0.09 & -0.09 & 1.00 & 1.00 & 0.79 & 0.75 \\
94 & 1.00 & 1.00 & 0.49 & 0.44 & 0.96 & 1.00 & -0.02 & -0.05 & 0.98 & 1.00 & 0.73 & 0.72 \\
95 & 1.00 & 1.00 & 0.50 & 0.41 & 0.97 & 1.00 & -0.05 & -0.08 & 0.99 & 1.00 & 0.76 & 0.74 \\
96 & 1.00 & 1.00 & 0.48 & 0.26 & 0.90 & 1.00 & 0.04 & -0.10 & 0.99 & 1.00 & 0.80 & 0.71 \\
97 & 1.00 & 1.00 & 0.41 & 0.40 & 0.98 & 1.00 & -0.06 & -0.03 & 0.98 & 1.00 & 0.77 & 0.72 \\
98 & 1.00 & 1.00 & 0.48 & 0.28 & 0.98 & 1.00 & -0.07 & -0.06 & 1.00 & 1.00 & 0.81 & 0.72 \\
99 & 1.00 & 1.00 & 0.51 & 0.34 & 0.99 & 1.00 & -0.11 & 0.02 & 0.99 & 1.00 & 0.77 & 0.74 \\
100 & 0.98 & 0.99 & 0.26 & 0.05 & 0.97 & 0.99 & 0.06 & 0.01 & 0.97 & 0.99 & 0.23 & 0.57 \\
101 & 0.97 & 0.99 & 0.34 & -0.01 & 0.94 & 1.00 & 0.01 & 0.05 & 0.98 & 1.00 & 0.48 & 0.62 \\
102 & 0.99 & 1.00 & 0.26 & 0.23 & 0.98 & 0.98 & -0.04 & -0.09 & 0.99 & 1.00 & 0.63 & 0.70 \\
103 & 0.99 & 1.00 & 0.22 & 0.40 & 0.90 & 1.00 & 0.02 & 0.02 & 0.91 & 1.00 & 0.33 & 0.57 \\
104 & 0.93 & 0.90 & 0.34 & 0.43 & 0.82 & 0.94 & 0.02 & 0.04 & 0.91 & 0.97 & -0.09 & -0.01 \\
105 & 0.97 & 0.97 & 0.02 & 0.09 & 0.89 & 0.91 & -0.00 & -0.02 & 0.88 & 0.97 & 0.03 & 0.05 \\
106 & 0.97 & 0.94 & 0.25 & 0.18 & 0.88 & 0.93 & 0.03 & -0.00 & 0.98 & 0.99 & -0.01 & 0.18 \\
107 & 0.98 & 0.99 & -0.05 & 0.17 & 0.97 & 0.98 & 0.02 & -0.01 & 0.99 & 0.99 & 0.26 & 0.15 \\
108 & 0.98 & 0.95 & 0.22 & 0.18 & 0.95 & 0.99 & -0.01 & 0.00 & 0.98 & 0.99 & 0.18 & 0.41 \\
109 & 0.99 & 0.99 & 0.36 & 0.33 & 0.97 & 0.94 & 0.02 & -0.01 & 0.99 & 1.00 & 0.35 & 0.52 \\
110 & 0.99 & 1.00 & 0.46 & 0.03 & 0.96 & 1.00 & 0.04 & -0.06 & 0.99 & 0.93 & 0.72 & 0.63 \\
111 & 1.00 & 1.00 & 0.30 & 0.29 & 0.93 & 1.00 & 0.08 & 0.03 & 0.99 & 1.00 & 0.30 & 0.30 \\
112 & 0.99 & 1.00 & 0.21 & -0.13 & 0.97 & 1.00 & 0.01 & 0.00 & 0.99 & 1.00 & 0.18 & 0.32 \\
113 & 1.00 & 1.00 & 0.45 & 0.01 & 0.83 & 1.00 & 0.04 & 0.08 & 0.96 & 1.00 & 0.18 & 0.62 \\
114 & 0.99 & 0.99 & 0.30 & 0.28 & 0.92 & 0.97 & 0.01 & -0.02 & 0.93 & 0.81 & 0.19 & -0.00 \\
115 & 0.99 & 1.00 & 0.46 & 0.34 & 0.91 & 1.00 & -0.00 & 0.08 & 0.97 & 1.00 & 0.30 & 0.48 \\
116 & 0.99 & 1.00 & 0.36 & 0.27 & 0.94 & 1.00 & -0.01 & 0.03 & 0.99 & 1.00 & 0.53 & 0.55 \\
117 & 0.99 & 1.00 & 0.39 & 0.24 & 0.93 & 1.00 & 0.06 & 0.02 & 0.93 & 1.00 & 0.71 & 0.72 \\
118 & 1.00 & 1.00 & 0.45 & 0.41 & 0.88 & 0.87 & -0.02 & -0.03 & 0.97 & 1.00 & 0.48 & 0.82 \\
119 & 1.00 & 1.00 & 0.45 & 0.32 & 0.87 & 0.90 & -0.02 & 0.02 & 0.93 & 1.00 & 0.13 & 0.74 \\
120 & 0.98 & 0.97 & 0.07 & 0.02 & 0.92 & 0.99 & 0.02 & 0.03 & 0.99 & 0.98 & 0.31 & 0.35 \\
\end{longtable}
\end{footnotesize}

\begin{table}[H]
\centering
\small
\caption{Game configuration parameters for each GLEE family.}
\label{tab:params}
\begin{tabular}{@{}lllp{6cm}@{}}
\toprule
\textbf{Family} & \textbf{Parameter} & \textbf{Values} & \textbf{Description} \\
\midrule
Bargaining & Stakes & \$100, \$10K, \$1M & Total amount to divide \\
           & Information & Complete, Incomplete & Whether players know each other's discount factors \\
           & Messages & Allowed, Not allowed & Whether free-text messages accompany offers \\
           & Discount $\delta_1$ & 0.8, 0.9, 0.95, 1.0 & Alice's per-round discount factor \\
           & Discount $\delta_2$ & 0.8, 0.9, 0.95, 1.0 & Bob's per-round discount factor \\
           & Max rounds & 12, $\infty$ & Maximum number of alternating offers \\
\midrule
Persuasion & Quality prob.\ $p$ & 0.33, 0.5, 0.8 & Prior probability of high quality \\
           & Value $v$ & 1.2, 1.25, 2.0, 3.0, 4.0 & High-quality value differential \\
           & Seller knowledge & Knows, Uninformed & Whether seller knows product quality \\
           & Buyer myopic & Yes, No & Single-round vs.\ multi-round buyer \\
           & Message type & Text, Binary & Seller's communication format \\
           & Price & \$100, \$10K, \$1M & Product price \\
\midrule
Negotiation & Price & \$100, \$10K, \$1M & Product base price \\
            & Information & Complete, Incomplete & Whether players know each other's valuations \\
            & Messages & Allowed, Not allowed & Free-text messages with offers \\
            & Max rounds & 10, 30 & Maximum negotiation rounds \\
            & Buyer value & 0.8, 1.0, 1.2, 1.5 & Buyer's private valuation multiplier ($V_B = \text{value} \times \text{price}$) \\
            & Seller value & 0.8, 1.0, 1.2, 1.5 & Seller's private valuation multiplier ($V_A = \text{value} \times \text{price}$) \\
\bottomrule
\end{tabular}
\end{table}

\begin{table}[H]
\centering
\small
\caption{Bargaining: base vs.\ aligned by configuration parameter. $N$: valid pairs after filtering; Filt.: excluded pairs; $p$: one-sided binomial.}
\label{tab:config_barg}
\begin{tabular}{@{}llrrrrl@{}}
\toprule
\textbf{Parameter} & \textbf{Value} & \textbf{$N$} & \textbf{Filt.} & \textbf{Base} & \textbf{Al.} & \textbf{$p$} \\
\midrule
Stakes          & \$100       & 74  & 46  & 73 & 1  & $4.0\!\times\!10^{-21}$ \\
                & \$10K       & 73  & 47  & 69 & 4  & $1.2\!\times\!10^{-16}$ \\
                & \$1M        & 64  & 56  & 62 & 2  & $1.1\!\times\!10^{-16}$ \\
Information     & Complete    & 76  & 44  & 72 & 4  & $1.8\!\times\!10^{-17}$ \\
                & Incomplete  & 78  & 42  & 75 & 3  & $2.6\!\times\!10^{-19}$ \\
Messages        & Allowed     & 66  & 54  & 64 & 2  & $3.0\!\times\!10^{-17}$ \\
                & Not allowed & 83  & 37  & 76 & 7  & $4.7\!\times\!10^{-16}$ \\
Discount $\delta_1$ & 0.8     & 17  & 103 & 10 & 7  & $0.31$ \\
                & 0.9         & 73  & 47  & 65 & 8  & $1.6\!\times\!10^{-12}$ \\
                & 0.95        & 60  & 60  & 57 & 3  & $3.1\!\times\!10^{-14}$ \\
                & 1.0         & 88  & 32  & 85 & 3  & $3.7\!\times\!10^{-22}$ \\
Discount $\delta_2$ & 0.8     & 80  & 40  & 76 & 4  & $1.4\!\times\!10^{-18}$ \\
                & 0.9         & 42  & 78  & 40 & 2  & $2.1\!\times\!10^{-10}$ \\
                & 0.95        & 78  & 42  & 74 & 4  & $5.0\!\times\!10^{-18}$ \\
                & 1.0         & 82  & 38  & 73 & 9  & $6.9\!\times\!10^{-14}$ \\
Round filter    & $= 1$       & 93  & 27  & 32 & 61 & $1.7\!\times\!10^{-3}$ \\
                & $\geq 2$    & 86  & 34  & 82 & 4  & $2.9\!\times\!10^{-20}$ \\
\bottomrule
\end{tabular}
\end{table}

\begin{table}[H]
\centering
\small
\caption{Persuasion: base vs.\ aligned by configuration parameter. $N$: valid pairs; Filt.: excluded; $p$: one-sided binomial.}
\label{tab:config_pers}
\begin{tabular}{@{}llrrrrl@{}}
\toprule
\textbf{Parameter} & \textbf{Value} & \textbf{$N$} & \textbf{Filt.} & \textbf{Base} & \textbf{Al.} & \textbf{$p$} \\
\midrule
Quality prob.\ $p$ & 0.33     & 51  & 69  & 33 & 18 & $0.024$ \\
                & 0.50        & 4   & 116 & 4  & 0  & $0.063$ \\
                & 0.80        & 0   & 120 & --& --& -- \\
Value $v$       & 1.2         & 35  & 85  & 31 & 4  & $1.7\!\times\!10^{-6}$ \\
                & 1.25        & 23  & 97  & 16 & 7  & $0.047$ \\
                & 2.0         & 37  & 83  & 33 & 4  & $5.4\!\times\!10^{-7}$ \\
                & 3.0         & 6   & 114 & 5  & 1  & $0.11$ \\
                & 4.0         & 15  & 105 & 12 & 3  & $0.018$ \\
Seller knowledge & Knows      & 31  & 89  & 29 & 2  & $2.3\!\times\!10^{-7}$ \\
                & Uninformed  & 46  & 74  & 32 & 14 & $5.7\!\times\!10^{-3}$ \\
Buyer myopic    & Yes         & 0   & 120 & --& --& -- \\
                & No          & 36  & 84  & 32 & 4  & $9.7\!\times\!10^{-7}$ \\
Message type    & Text        & 33  & 87  & 32 & 1  & $4.0\!\times\!10^{-9}$ \\
                & Binary      & 47  & 73  & 43 & 4  & $1.4\!\times\!10^{-9}$ \\
Price           & \$100       & 32  & 88  & 26 & 6  & $2.7\!\times\!10^{-4}$ \\
                & \$10K       & 34  & 86  & 34 & 0  & $5.8\!\times\!10^{-11}$ \\
                & \$1M        & 0   & 120 & --& --& -- \\
Round filter    & $= 1$       & 53  & 67  & 23 & 30 & $0.21$ \\
                & $\geq 2$    & 39  & 81  & 31 & 8  & $1.5\!\times\!10^{-4}$ \\
\bottomrule
\end{tabular}
\end{table}

\begin{table}[H]
\centering
\small
\caption{Negotiation: base vs.\ aligned by configuration parameter. $N$: valid pairs; Filt.: excluded; $p$: one-sided binomial.}
\label{tab:config_nego}
\begin{tabular}{@{}llrrrrl@{}}
\toprule
\textbf{Parameter} & \textbf{Value} & \textbf{$N$} & \textbf{Filt.} & \textbf{Base} & \textbf{Al.} & \textbf{$p$} \\
\midrule
Information     & Complete    & 21  & 99  & 20 & 1  & $1.1\!\times\!10^{-5}$ \\
                & Incomplete  & 40  & 80  & 39 & 1  & $3.7\!\times\!10^{-11}$ \\
Messages        & Allowed     & 28  & 92  & 27 & 1  & $1.1\!\times\!10^{-7}$ \\
                & Not allowed & 32  & 88  & 31 & 1  & $7.7\!\times\!10^{-9}$ \\
Max rounds      & 10          & 55  & 65  & 54 & 1  & $1.6\!\times\!10^{-15}$ \\
                & 30          & 7   & 113 & 6  & 1  & $0.063$ \\
Price           & \$100       & 29  & 91  & 27 & 2  & $8.1\!\times\!10^{-7}$ \\
                & \$10K       & 20  & 100 & 19 & 1  & $2.0\!\times\!10^{-5}$ \\
                & \$1M        & 44  & 76  & 42 & 2  & $5.6\!\times\!10^{-11}$ \\
Value asymmetry & Buyer $>$ Seller & 28 & 92 & 26 & 2 & $1.5\!\times\!10^{-6}$ \\
                & Seller $>$ Buyer & 38 & 82 & 36 & 2 & $2.7\!\times\!10^{-9}$ \\
                & Equal       & 27  & 93  & 27 & 0  & $7.5\!\times\!10^{-9}$ \\
Round filter    & $= 1$       & 72  & 48  & 33 & 39 & $0.28$ \\
                & $\geq 2$    & 57  & 63  & 56 & 1  & $4.0\!\times\!10^{-16}$ \\
\bottomrule
\end{tabular}
\end{table}

\begin{table}[H]
\centering
\small
\caption{Matrix games (PD and BoS): base vs.\ aligned by round phase. $N$: valid pairs; Filt.: excluded; $p$: one-sided binomial.}
\label{tab:config_matrix}
\begin{tabular}{@{}llrrrrl@{}}
\toprule
\textbf{Game} & \textbf{Round Phase} & \textbf{$N$} & \textbf{Filt.} & \textbf{Base} & \textbf{Al.} & \textbf{$p$} \\
\midrule
PD  & Early (1--3)  & 32  & 88  & 12 & 20 & $0.11$ \\
    & Mid (4--7)    & 116 & 4   & 93 & 23 & $1.8\!\times\!10^{-11}$ \\
    & Late (8--10)  & 119 & 1   & 92 & 27 & $8.7\!\times\!10^{-10}$ \\
\midrule
BoS & Early (1--3)  & 0   & 120 & --& --& -- \\
    & Mid (4--7)    & 11  & 109 & 10 & 1  & $5.9\!\times\!10^{-3}$ \\
    & Late (8--10)  & 111 & 9   & 93 & 18 & $1.1\!\times\!10^{-13}$ \\
\bottomrule
\end{tabular}
\end{table}

\end{document}